\documentclass[10pt,twocolumn,letterpaper]{article}

\usepackage[cvprfinal]{cvpr}      
\usepackage[pagebackref,breaklinks,colorlinks]{hyperref}

\usepackage{multirow,epsfig,pdflscape,graphicx,amsmath,amssymb}
\usepackage{float}
\usepackage{amsmath,amssymb,amsfonts}
\usepackage{textcomp}
\usepackage{mathrsfs}
\usepackage{empheq}
\usepackage{rotating}
\usepackage{sidecap, caption}
\usepackage{color, colortbl}
\definecolor{LightCyan}{rgb}{0.88,1,1}
\usepackage[table]{xcolor}
\usepackage{url}            % simple URL typesetting
\usepackage{booktabs}       % professional-quality tables
\usepackage{nicefrac}       % compact symbols for 1/2, etc.
\usepackage{microtype}      % microtypography
\usepackage{lipsum}
\usepackage{cite,multirow}
\usepackage{array, caption, tabularx,  ragged2e,  booktabs}
\usepackage{longtable}
\usepackage{textcomp}
\usepackage{siunitx}
\usepackage{array}
\usepackage[linesnumbered,ruled,vlined]{algorithm2e}
\usepackage{algpseudocode}
\usepackage{romannum}
\usepackage{graphicx}
\usepackage{xcolor,comment}

\usepackage{times}
\usepackage{epsfig}
\usepackage{amsmath}
\usepackage{amssymb}
\usepackage{appendix}
\usepackage{etoolbox}
\usepackage{soul}
\usepackage{tabularx}
\usepackage{caption}
\usepackage{subcaption,xspace}

\usepackage[capitalize]{cleveref}

\usepackage[inline]{enumitem}

\usepackage[capitalize]{cleveref}
\crefname{section}{Sec.}{Secs.}
\Crefname{section}{Section}{Sections}
\Crefname{table}{Table}{Tables}
\crefname{table}{Tab.}{Tabs.}

\BeforeBeginEnvironment{appendices}{\clearpage}

\newcolumntype{Y}{>{\centering\arraybackslash}X}

\newcommand{\argmax}[1]{\underset{#1}{\operatorname{arg}\,\operatorname{max}}\;}

\usepackage{graphicx} 
\usepackage{caption}

\def\X{{\mathcal{X}}}
\def\Y{{\mathcal{Y}}}
\def\D{{\mathcal{D}}}
\def\L{{\mathcal{L}}}

\def\etal{{et al.\xspace}}
\def\ood{{OOD\xspace}}

\newcolumntype{M}[1]{>{\centering\arraybackslash}m{#1}}
\newcolumntype{N}{@{}m{0pt}@{}@{}}

\usepackage{enumitem}

\newcommand{\myfirstpara}[1]{\noindent \textbf{#1:}}
\newcommand{\mypara}[1]{\vspace{0.2em} \myfirstpara{#1}}

\graphicspath{{images/}}
\graphicspath{{./imgs/}}

\def\nll{\textsc{nll}\xspace}
\def\nllns{\textsc{nll}}
\def\cnn{\textsc{cnn}\xspace}
\def\dnn{\textsc{dnn}\xspace}
\def\dnns{\textsc{dnn}s\xspace}
\def\cams{\textsc{CAM}s\xspace}
\def\sota{\textsc{sota}\xspace}
\def\cifar{\textsc{cifar}\xspace}
\def\svhn{\textsc{svhn}\xspace}
\def\pacs{\textsc{pacs}\xspace}
\def\mdca{\textsc{mdca}\xspace}
\def\sce{\textsc{sce}\xspace}
\def\ece{\textsc{ece}\xspace}
\def\pascal{\textsc{pascal-voc}\xspace}
\def\mnist{\textsc{mnist}\xspace}
\def\ts{\textsc{ts}\xspace}
\def\ood{\textsc{ood}\xspace}
\def\dca{\textsc{dca}\xspace}
\def\mmce{\textsc{mmce}\xspace}
\def\dc{\textsc{dc}\xspace}
\def\ls{\textsc{ls}\xspace}
\def\mce{\textsc{mce}\xspace}
\def\bs{\textsc{bs}\xspace}
\def\fl{\textsc{fl}\xspace}
\def\flns{\textsc{fl}}
\def\flsd{\textsc{flsd}\xspace}
\def\te{\textsc{te}\xspace}

\def\sbf{\mathbf{s}}
\def\xbf{\mathbf{x}}

\def\yh{\widehat{y}}
\def\sh{\widehat{s}}

\begin{document}

\title{A Stitch in Time Saves Nine: \\ A Train-Time Regularizing Loss for Improved Neural Network Calibration \vspace{-0.5em}}

\author{
	Ramya Hebbalaguppe$^{1,2}$\textsuperscript{\textsection} 
	\quad Jatin Prakash$^{1}$\textsuperscript{\textsection} 
	\quad Neelabh Madan$^{1}$\textsuperscript{\textsection} 
	\quad Chetan Arora$^1$ 
	\\ 
	$^1$Indian Institute of Technology Delhi, India 
	\quad$^2$TCS Research, India
	\\
	%{\tt\small\{anz198720, cs1180344, me1180698, chetan\}@iitd.ac.in}
	{\small \url{https://github.com/mdca-loss}}
}

\twocolumn[{%
	\renewcommand\twocolumn[1][]{#1}%
	\maketitle
	\begin{center}
		\vspace{-1em}
		\includegraphics[width=0.9\linewidth]{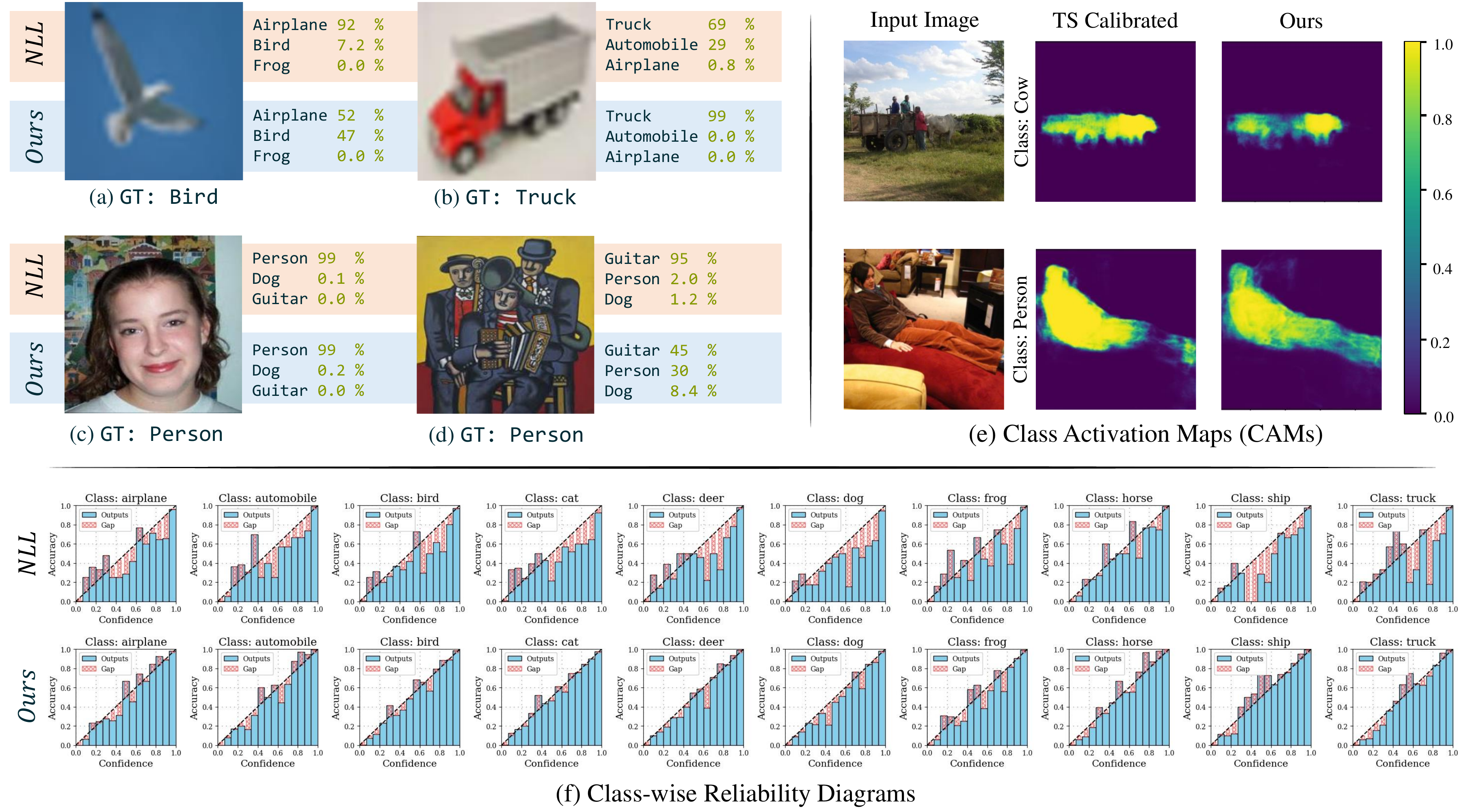}
		\vspace{-0.5em}
		\captionof{figure}{
			We present a new regularizing loss (\mdca) for train time calibration of Deep Neural Networks (\dnn). Figures (a)-(d) shows comparison with a model trained using Cross Entropy loss (\nll), and ours (\nll+\mdca). In (a), a \dnn trained using \nll makes an incorrect but over-confident prediction. Whereas, training with \mdca reduces the confidence of the mis-predicted label, and increases confidence of the second-highest confident, but correct label. In (b) for a \cifar10 minority class, ``truck'', a model trained with \mdca confidently predicts the correct label as compared to a \nll trained model. In (c) and (d), we show an image from in-domain and out-of-domain dataset. In (c), picture is taken from ``Photo'' domain (in-domain) of the \pacs \cite{pacspaper} dataset on which we trained the \dnn. Both the models trained with \mdca and \nll predict high confidence score for the correct label. However, in (d), when we change the domain to ``Art'' (out-of-domain), we see \nll trained model makes highly over-confident mistake on domain shift, whereas, \mdca trained model remains calibrated. In (e) we show the Class Activation Maps (\cams) for a model calibrated with Temperature Scaling (\ts), and ours for label cow (top row), and person (bottom row). More accurate \cams show that training with \mdca improves model explainability. (f) shows class-wise reliability diagrams of models trained with \nll and \mdca. \mdca leads to models which are calibrated for all classes.}
		\label{fig:teaser}
	\end{center}%
}]
\begingroup\renewcommand\thefootnote{\textsection}
\footnotetext{Equal contribution}
\endgroup

\begin{abstract}
\vspace{-0.5em}
Deep Neural Networks (\dnns) are known to make overconfident mistakes, which makes their use problematic in safety-critical applications. State-of-the-art (\sota) calibration techniques improve on the confidence of predicted labels alone, and leave the confidence of non-max classes (e.g. top-2, top-5) uncalibrated. Such calibration is not suitable for label refinement using post-processing. Further, most \sota techniques learn a few hyper-parameters post-hoc, leaving out the scope for image, or pixel specific calibration. This makes them unsuitable for calibration under domain shift, or for dense prediction tasks like semantic segmentation. In this paper, we argue for intervening at the train time itself, so as to directly produce calibrated \dnn models. We propose a novel auxiliary loss function: \textbf{M}ulti-class \textbf{D}ifference in \textbf{C}onfidence and \textbf{A}ccuracy (\mdca), to achieve the same. \mdca can be used in conjunction with other application/task specific loss functions. We show that training with \mdca leads to better calibrated models in terms of Expected Calibration Error (\ece), and Static Calibration Error (\sce) on image classification, and segmentation tasks. We report \ece(\sce) score of 0.72 (1.60) on the \cifar100 dataset, in comparison to 1.90 (1.71) by the \sota. Under domain shift, a ResNet-18 model trained on \pacs dataset using \mdca gives a average \ece(\sce) score of 19.7 (9.7) across all domains, compared to 24.2 (11.8) by the \sota. For segmentation task, we report a $2\times$ reduction in calibration error on \pascal dataset in comparison to Focal Loss \cite{ogfocalloss}. Finally, \mdca training improves calibration even on imbalanced data, and for natural language classification tasks.
\end{abstract}
\vspace{-1em}
\section{Introduction}

Deep Neural Networks (\dnns) have shown promising results for various pattern recognition tasks in recent years. In a classification setting, with input $\xbf \in \mathcal{X}$, and label $y \in \Y = \{1, \ldots, K\}$, a \dnn typically outputs a \emph{confidence} score vector $\sbf \in \mathbb{R}^K$. The vector, $\sbf$, is also a valid probability vector, and each element of $\sbf$ is assumed to be the predicted confidence for the corresponding label. It has been shown in recent years that the confidence vector, $\sbf$, output by a \dnn is often poorly calibrated \cite{guo2017calibration, minderer2021revisiting}. That is:
\begin{equation}
\mathbb{P} \Big( \yh=y^* ~ \big\vert ~ \sbf[\yh] \Big) ~ \neq ~ \sbf[\yh],
\end{equation}
where $\widehat{y}$, and $y^*$ are the predicted, and true label respectively for a sample. E.g. if a \dnn predicts a class ``truck'' for an image with score 0.7, then a network is calibrated, if the probability that the image actually contains a truck is 0.7. If the probability is lower, a network is said to be over-confident, and under-confident if probability is higher. For a pixel-wise prediction task like semantic segmentation, we would like to calibrate prediction for each pixel. Similarly, we would like calibration to hold not only for the predicted label, i.e. $\yh = \argmax{y \in \Y} \sbf[y]$, but for the whole vector $\sbf$ (all labels), i.e., $\forall y \in \Y$. 

One of the main reasons for the miscalibration is the specific training regimen used. Most modern \dnns, when trained for classification in a supervised learning setting, are trained using one-hot encoding that have all the probability mass centered in one class; the training labels are thus zero-entropy signals that admit no uncertainty about the input \cite{thulasidasan2019mixup}. The \dnn is thus trained to become overconfident. Besides creating a general distrust in the model predictions, the miscalibration is especially problematic in safety critical applications, such as self-driving cars \cite{grigorescu2020survey}, legal research \cite{yu2019s} and healthcare \cite{chromosome, Dusenberry_2020}, where giving the correct confidence for a predicted label is as important as the correct label prediction itself.

Researchers have tried to address miscalibration by learning a post-hoc transformation of the output vector so that the confidence of the predicted label matches with the likelihood of the label for the sample \cite{hinton2015distilling,DBLP:journals/corr/HendrycksG16c}. Since such techniques focus on the predicted label only, they could end up calibrating only the label which has maximum confidence for each sample. Hence, in a multi-class setting, the labels with non-maximal confidence scores remain uncalibrated. This makes any post-processing for label refinement, such as posterior inference using MRF-MAP \cite{deeplabv1}, ineffective. 

In this paper we argue for the calibration at the train-time. Unlike post-hoc calibration techniques that use limited parameters\footnote{For example, Temperature scaling (\ts) calibrates uses a single global scalar, $T$; and Dirichlet Calibration (\texttt{DC}) uses $\mathcal{O}(K^2)$ hyper-parameters for $K$ classes to calibrate the model output}, a train time strategy allows exploiting millions of learnable parameters of DNN itself, thus providing a flexible learning more suited to image and pixel specific transformation for model calibration. Our experiments under domain shift, and for a dense predict task (semantic segmentation) shows the strength of the approach.
% A train time intervention is not required to restrict itself to maintaining the rank, as additional loss function to control the accuracy can always be added. Further, instead of requiring a few separate hyper-parameters for calibration, the whole weight vector of a \dnn comprising of the network parameters now becomes available for image, and pixel specific calibration. Beside, such intervention also enables calibrating the whole confidence vector.
%most calibration techniques learn a few image agnostic parameters in a post-hoc fashion such that rank of the predicted label does not change. Maintaining the rank is important because there is no arrangement in these techniques to guard against the accuracy drop if the rank changes. For example, temperature scaling (\ts) for classification probability calibration determines only one global scaling constant (temperature), and does not capture spatial miscalibration changes in images. Besides, the non-max labels remain miscalibrated, and such models easily lose their calibration under domain shift.

%In this paper we argue for the intervention at the train time itself. A train time intervention is not required to restrict itself to maintaining the rank, as additional loss function to control the accuracy can always be added. Further, instead of requiring a few separate hyper-parameters for calibration, the whole weight vector of a \dnn comprising of the network parameters now becomes available for image, and pixel specific calibration. Beside, such intervention also enables calibrating the whole confidence vector.

Armed with the above insight, we propose a novel auxiliary loss function: \textbf{M}ulti-class \textbf{D}ifference in \textbf{C}onfidence and \textbf{A}ccuracy (\mdca). The proposed loss function is designed to be used during the training stage in conjunction with other application specific loss functions, and overcomes the non-differentiablity of the loss functions proposed in earlier methods. Though we do not advocate it, the proposed technique is complimentary to the post-hoc techniques which may still be used after the training, if there is a separate hold-out dataset available for exploitation. Since ours is a train time calibration approach, it implies good regularization for the predictions. We show that models trained using our loss
function remain calibrated even under domain shift.

\mypara{Contributions} We make the following key contributions:
\begin{enumerate*}[label=\textbf{(\arabic*)}]%[leftmargin=*, topsep=0.2em, itemsep=-0.2em]
\item A trainable \dnn calibration method with inclusion of a novel auxiliary loss function, termed \mdca, that takes into account the entire confidence vector in a multi-class setting. Our loss function is differentiable and can be used in conjunction with any existing loss term. We show experiments with Cross-Entropy, Label Smoothing \cite{labelsmoothinghelp}, and Focal Loss \cite{ogfocalloss}. 
%
% \item Our loss function is differentiable. This makes latest gradient descent techniques applicable for optimizing our loss function, leading to better minimization of the objective function, and improved calibration.
%
\item Our approach is on par with post-hoc methods \cite{Dirichlet,guo2017calibration} without the need for hold-out set making the deployment more practical (See \cref{tab:TrainVsPH}).
\item Our loss function is a powerful regularizer, maintaining calibration even under domain/dataset drift and dataset imbalance which We demonstrate on \pacs \cite{pacspaper}, Rotated MNIST \cite{lecun2010mnist} and imbalanced \cifar 10 datasets.
\item Although the focus is primarily on image classification, our experiments on multi-class semantic segmentation show that our technique outperforms \ts based calibration, and Focal Loss \cite{ogfocalloss}. We also show the effectiveness of our approach on natural language classification task on 20Newsgroup dataset~\cite{Lang95}. 
\end{enumerate*}
%To support reproducible research, we release complete source-code of our work \footnote{\small \url{https://github.com/mdca-aux-loss/MDCA-Calibration}}. 

\section{Related Work}
\label{sec:relWorks}
Techniques for calibrating \dnns can be broadly classified into train-time calibration, post-hoc calibration, and calibration through Out-Of-Distribution (\ood). Train-time calibration integrate model calibration during the training procedure while a post-hoc calibration method utilizes a hold-out set to tune the calibration measures. On the other hand,learning to reject \ood samples (at train-time or post-hoc) mitigates overconfidence and thus, calibrates \dnns.

\mypara{Train-Time Calibration}  
One of the earliest train-time methods proposes Brier Score for the calibrating binary probabilistic forecast \cite{brierloss}. \cite{guo2017calibration} show models trained with Negative-Log-Likelihood (\nll) tend to be over-confident and empirically show a disconnect between \nll and accuracy. Specifically, the overconfident scores necessitates re-calibration. A common calibration approach is to use additional loss terms other than the \nll loss: \cite{pereyra2017regularizing} use entropy as a regularization term whereas M{\"u}ller \etal \cite{labelsmoothinghelp} propose Label Smoothing (\ls) \cite{originallabelsmoothing} on soft-targets which aids in improving calibration. Recently, \cite{focallosspaper} showed that focal loss \cite{ogfocalloss} can implicitly calibrate \dnns by reducing the KL-divergence between predicted and target distribution whilst increasing the entropy of the predicted distribution, thereby preventing the model from becoming overconfident. Liang \etal \cite{dcapaper} have proposed an auxiliary loss term, \dca, which is added with Cross-Entropy to help calibrate the model. The \dca term penalizes the model when the cross-entropy loss is reduced, but the accuracy remains the same, i.e., when the over-fitting occurs. \cite{kumarpaper} propose to use \mmce, an auxiliary loss term for calibration, computed using a reproducing kernel in a Hilbert space~\cite{rkhskernel}. Maro{\~n}as \etal \cite{maronas2021calibration} analyse MixUp \cite{mixup-augmentation} data augmentation for calibrating DNNs and conclude Mixup does not necessarily improve calibration.

\mypara{Post-Hoc Calibration} 
Post-hoc calibration techniques calibrate a model using a hold-out training set, which is usually the validation set. Temperature scaling (\ts) smoothes the logits to calibrate a \dnn. Specifically, \ts is a variant of Platt scaling \cite{platt1999probabilistic} that works by dividing the logits by a scalar $T > 0$,  learnt on a hold-out training set, prior to taking a softmax. The downside of using \ts during calibration is reduction in confidence of every prediction, including the correct one. A more general version of \ts transforms the logits using a matrix scaling. The matrix $M$ is learnt using the hold-out set similar to \ts. Dirichlet calibration~(\dc) uses Dirichlet distributions to extend the Beta-calibration \cite{beta-cal-paper} method for binary classification to a multi-class one. \dc is easy to implement as an extra layer in a neural network on log-transformed class probabilities, which is learnt on a hold-out set. Meta-calibration propose differentiable ECE-driven  calibration to obtain well-calibrated and high-accuracy models \cite{bohdal2021meta}. Islam \etal \cite{islamclass} propose class-distribution-aware \ts and \ls that can be used as a post-hoc calibration. They use a class-distribution aware vector for \ts/\ls to fix the overconfidence. Ding \etal \cite{local-TS} propose a spatially localized calibration approach for semantic segmentation.

\mypara{Calibration Through \ood Detection} 
Hein \etal \cite{hein2019relu} show that one of the main reasons behind the overconfidence in \dnns is the usage of ReLu activation that gives high confidence predictions when the input sample is far away from the training data. They propose data augmentation using adversarial training, which enforces low confidence predictions for samples far away from the training data. Guo \etal \cite{guo2017calibration} analyze the effect of width, and depth of a \dnn, batch normalization, and weight decay on the calibration. Karimi \etal \cite{ood-spectral} use spectral analysis on initial layers of a \cnn to determine \ood sample and calibrate the \dnn. We refer the reader to \cite{hendrycks2018deep, devries2018learning, padhy2020revisiting, meronen2020stationary} for other representative works on calibrating a \dnn through \ood detection.

%Our work focuses on calibrating the entire confidence vector at train-time, and does not need a hold-out dataset to tune the hyper-parameters. The proposed loss function can be used in conjunction with other application specific ones. We demonstrate the efficacy of training with our loss-function in a stand-alone manner, as well as in conjunction with other commonly used loss functions, and state-of-the-art post-hoc techniques discussed above. 

\section{Proposed Methodology}
\label{sec:Propmethodology}

\mypara{Calibration}
A calibrated classifier outputs confidence scores that matches the empirical frequency of correctness. If a calibrated model predicts an event with $0.7$ confidence, then $70\%$ of the times the event transpires. If the empirical occurrence of the event is $<70\%$ then the model is overconfident, and if the empirical probability $>70\%$ then the model is under-confident. Formally, we define calibration in a classical supervised setting as follows. Let $\D=\langle(x_i,y_i)\rangle_{i=1}^N$ denote a dataset consisting of $N$ samples from a joint distribution $\D(\X,\Y)$, where for each sample $x_i \in \X$ is the input and $y_i^* \in \Y = \{1, 2, ..., K\}$ is the ground-truth class label. Let $\sbf \in \mathbb{R}^K$, and $\sbf_i[y] = f_\theta(x_i)$ be the confidence that a \dnn, $f$, with model parameters $\theta$ predicts for a class $y$ on a given input $x_i$. The class, $\yh_i$, predicted by $f$ for a sample $x_i$ is computed as: 
\begin{equation}
    \yh_i = \argmax{y \in \Y}  \sbf_i[y].
\end{equation}
The confidence for the predicted class is correspondingly computed as $\sh_i = \max_{y\in \Y} s_i[y]$. A model is said to be \emph{perfectly calibrated} \cite{guo2017calibration} when, for each sample $(x, y) \in \D$: 
\begin{equation}
    \mathbb{P}(y = y^* ~ | ~ \sbf[y] = s ) = s.
    \label{equ:calib2}
\end{equation}
Note that the perfect calibration requires each score value (and not only the $\widehat{s}$) to be calibrated. On the other hand, most calibration techniques focus only on the predicted class. That is, they only ensure that: $\mathbb{P}(\widehat{y}_i = y_i^* ~ | ~ \sh_i) = \sh_i$. 

\mypara{Expected Calibration Error (\ece)}
\ece is calculated by computing a weighted average of the differences in the confidence of the predicted class, and the accuracy of the samples, predicted with a particular confidence score \cite{ecepaper}:
\begin{equation}
    \ece = \sum_{i=1}^M \frac{B_i}{N} \Big\lvert A_i - C_i \Big\rvert.
    \label{equ:ece}
\end{equation}
Here $N$ is the total number of samples, and the weighting is done on the basis of the fraction of samples in a given confidence bin/interval. Since the confidence values are in a continuous interval, for the computation of \ece, we divide the confidence range $[0,1]$ into $M$ equidistant bins, where $i^\text{th}$ bin is the interval $(\frac{i-1}{M}, \frac{i}{M}]$ in the confidence range, and $B_i$, represents the number of samples in the $i^\text{th}$ bin. Further, $A_i =\frac{1}{|B_i|}\sum_{j\in B_i} \mathbb{I} (\hat{y}_j=y_j)$, denotes accuracy for the samples in bin $B_i$, and $C_i=\frac{1}{|B_i|}\sum_{j: \sh_j \in B_i} \sh_j$, is the average predicted confidence of the samples, such that $\sh_j \in B_i$. The evaluation of \dnn calibration via \ece suffers from the following shortcomings: (a) \ece does not measure the calibration of all score values in the confidence vector, and (b) the metric is not differentiable, and hence can not be incorporated as a loss term during training procedure itself. Specifically, non-differentiablity arises due to binning samples into bins $B_i$.

\mypara{Maximum Calibration Error (\mce)}
\mce is defined as the maximum absolute difference between the average accuracy and average confidence of each bin:
\[
	\mce = \max_{i \in {1,...,M}} \big\lvert A_i - C_i \big\rvert. 
\]
The \textit{max} operator ends up pruning a lot of useful information about calibration, making the metric not-so-popular. However, it does represent a statistical value that can be used to discriminate large differences in calibration.

\mypara{Static Calibration Error (\sce)}
\sce is a recently proposed metric to measure calibration by \cite{nixon2019measuring}: 
\begin{equation}
    \sce = \frac{1}{K} \sum_{i=1}^M \sum_{j=1}^K \frac{B_{i,j}}{N} \big\lvert A_{i,j} - C_{i,j} \big\rvert,
    \label{equ:sce}
\end{equation}
where, $K$ denotes the number of classes, and $B_{i,j}$ denotes number of samples of the $j^\text{th}$ class in the $i^\text{th}$ bin. Further, $A_{i,j} = \frac{1}{B_{i,j}} \sum_{k \in B_{i,j}} \mathbb{I} (j = y_{k})$ is the accuracy for the samples of $j^\text{th}$ class in the $i^{th}$ bin, and $C_{i,j} = \frac{1}{B_{i,j}} \sum_{k \in B_{i,j}} \mathbf{s}_k[j]$ or average confidence for the $j^\text{th}$ class in the $i^\text{th}$ bin. \textbf{Classwise-\ece} \cite{Dirichlet} is another metric for measuring calibration in a multi-class setting, but is identical to Static Calibration Error (\sce). It is easy to see that \sce is a simple class-wise extension to \ece. Since \sce takes into account the whole confidence vector, it allows us to measure calibration of the non-predicted classes as well. Note that, similar to \ece, the metric \sce is also non-differentiable, and can not be used as a loss term during training. 

\mypara{Class-$j$-\ece} 
\cite{Dirichlet} has proposed to evaluate calibration error of each class independent of other classes. This allows one to capture the contribution of a single class $j$ to the overall \sce (or classwise-\ece) error. We refer to this metric as class-$j$-\ece in our results/discussion. 

\subsection{Proposed Auxiliary loss: MDCA}

We propose a novel multi-class calibration technique using the proposed auxiliary loss function. The loss function is inspired from SCE \cite{nixon2019measuring} but avoids the non-differentiability caused due to binning $B_{i,j}$ as shown in \cref{equ:sce} \cite{dcapaper}. Our calibration technique is \textbf{independent} of the binning scheme/bins. This is important, because as \cite{widmann2019calibration} and \cite{kumar2019verified} have also highlighted, binning scheme leads to underestimated calibration errors. We name our loss function, \textit{Multi-class Difference of Confidence and Accuracy (\mdca)}, and apply it for each \textbf{mini-batch} during training. The loss is defined as follows:
% We define MDCA loss for a mini-batch of training data as:
\begin{equation}
    \L_\mdca = \frac{1}{K} \sum_{j=1}^K \Big\lvert \frac{1}{N_b}\sum_{i=1}^M \sbf_i[j] - \frac{1}{N_b} \sum_{i=1}^M  q_{i}[j] \Big\rvert,
    \label{equ:mdca}
\end{equation}
%However, similar to $\ell_1$ norm, our loss function is fully differentiable (except at 0)
where $q_{i}[j]=1$ if label $j$ is the ground truth label for sample $i$, i.e. $j=y^*_i$, else $q_{i}[j]=0$. 
%($q$ is basically a ground-truth one-hot matrix). 
Note the second term inside $|\cdot|$ corresponds to average count of samples in a mini-batch containing $N_b$ training samples. Since the average count is a constant value so learning gradients solely depends on the first term representing confidence assigned by the \dnn. $K$ denotes number of classes. $\mathcal{L}_\mdca$ is computed on a mini-batch, and the modulus operation ($|\cdot|$) implies that the summations are not interchangeable
\footnote{Note that $\mathcal{L}_{\mdca}$ may appear similar to $\mathcal{L}_{1}$ loss due to the usage of the modulus in both. However, the two loss functions are very different. Mathematically, $\mathcal{L}_1 = \frac{1}{K\cdot {N_{b}}} \sum_{j=1}^K \sum_{i=1}^{N_{b}} \Big\lvert \textbf{s}_i[j] - q_{i}[j] \Big\rvert$ whereas $\L_\mdca$ is as given in \cref{equ:mdca}. The two terms inside the modulus of $\L_\mdca$ loss represent mean statistic for a particular class, $j$ (motivated by our objective of class-wise calibration), whereas, in the case of $\mathcal{L}_{1}$ the modulus operate on a single sample. %(In fact, $\mathcal{L}_{1} \geq \mathcal{L}_{\mdca}$)
}.
Further, $\sbf_i[j]$ represents the confidence score by a \dnn for the $j^\text{th}$ class, of $i^{th}$ sample in the mini-batch. 
% We refer the reader to to supplementary material for derivation of differentiablity of $\mathcal{L}_{MDCA}$.

Note that $\L_\mdca$ is differentiable, whereas, the loss given by \dca \cite{dcapaper} involves accuracy over the mini-batch, and is non-differentiable. % Instead, \cite{dcapaper} suggest to back-propagate the \ece based loss ?? 
The differentiablity of our loss function ensures that it can be easily used in conjunction with other application specific loss functions as follows:
\begin{equation}
\mathcal{L}_\text{total} = \mathcal{L}_{C} + \beta \cdot \mathcal{L}_\mdca,
 \label{equ:totalLoss}
\end{equation} 
where $\beta$ is a hyperparameter to control the relative importance with respect to application specific losses, and is typically found using a validation set.  $\mathcal{L}_{C}$ is a standard classification loss, such as Cross Entropy, Label Smoothing~\cite{originallabelsmoothing}, or Focal loss~\cite{ogfocalloss}. Our experiments indicate that the proposed \mdca loss in conjunction with focal loss gives best calibration performance. 

%\mypara{Rationale for using $\mathcal{L}_{MDCA}$ }
Ideally to achieve \emph{confidence calibration}, we want the average prediction confidence to be same as accuracy of the model. However, in \emph{multiclass calibration}, we want average prediction confidence of every class $k_i$ to match with its average occurrence in the data-distribution. In $\mathcal{L}_{MDCA}$, we explicitly capture this idea for every mini-batch i.e. we intuitively want that $\tilde{s}[k_i] \approx \tilde{q}[k_i]$ (where $\tilde{s}[k_i],\tilde{q}[k_i]$ is the average prediction confidence and the average count class $k_i$ in a mini-batch respectively). Any deviation from this leads DNN to be penalized by $\mathcal{L}_{MDCA}$.

\begin{table*}[h]
	\centering
	\resizebox{ \linewidth}{!}{%
		\begin{tabular}{cc||cc|cc||cc|cc||cc|cc}
			\toprule
			&
			&
			\multicolumn{2}{c|}{\textbf{NLL}} &
			\multicolumn{2}{c||}{\textbf{NLL+MDCA}} &
			\multicolumn{2}{c|}{\textbf{LS \cite{labelsmoothinghelp}}} &
			\multicolumn{2}{c||}{\textbf{LS+MDCA}} &
			\multicolumn{2}{c|}{\textbf{FL \cite{ogfocalloss}}} &
			\multicolumn{2}{c}{\textbf{FL+MDCA}} \\
			\multirow{-2}{*}{\textbf{Dataset}} &
			\multirow{-2}{*}{\textbf{Model}} &
			SCE(10$^{-3}$) &
			ECE (\%) &
			SCE(10$^{-3}$) &
			ECE (\%) &
			SCE(10$^{-3}$) &
			ECE (\%) &
			SCE(10$^{-3}$) &
			ECE (\%) &
			SCE(10$^{-3}$) &
			ECE (\%) &
			SCE(10$^{-3}$) &
			ECE (\%) \\
			\midrule
			\multirow{2}{*}{CIFAR10}  & ResNet32       & 8.68 & 4.25 &  4.63 & 1.69 & 14.08 & 6.28 &  10.39 & 4.31 & 4.60 & 1.76  &  \textbf{3.22} & \textbf{0.93}  \\
			& ResNet56       & 7.11 & 3.27 & 6.87 &	3.15 & 12.54 & 5.38 &  9.88 &	3.97 & 4.18 & 1.11  & \textbf{2.93} & \textbf{0.70}               \\ \midrule
			\multirow{2}{*}{CIFAR100} & ResNet32       & 3.03 & 12.45 & 2.59 & 9.94 & 1.99 & 2.09 &  1.74 & 	2.29 & 1.83 & 1.62  & \textbf{ 1.72} & 	\textbf{1.49}         \\
			& ResNet56       & 2.50 & 9.32 & 2.41 & 	8.95 & 1.73 & 8.94 &  1.68 & 	1.48 & 1.66 & 2.29  &  \textbf{1.60 }& \textbf{	0.72}    \\ \midrule
			\multirow{2}{*}{SVHN}     & ResNet20       & 3.43 & 1.64 & \textbf{1.46} & 	\textbf{0.43} & 18.80 & 8.88 &  13.91 & 	6.46 & 2.54 & 0.89  &  1.90 & 	0.47       \\
			& ResNet56       & 3.84 & 1.82 &  \textbf{1.47} & 	0.53 & 21.08 & 10.00 &  17.62 & 	8.43 & 7.85 & 3.89  &  1.51 & 	\textbf{0.23}         \\ \midrule
			Mendeley V2               & ResNet50       & 131.2 & 4.78 & 88.14	 & 3.63 & 103.8 & \textbf{2.68} &  97.38 & 	5.03 & 108.3 & 8.17  &  \textbf{85.68} & 	4.81      \\ \midrule
%			Kather5000                & ResNet34       & 31.70 & 7.14 & 33.95 & 	8.26 & 46.63 & 15.36 &  45.12 & 	15.13 & 41.69 & 13.62  &  41.39 & 	13.45                 \\ \midrule
			Tiny-ImageNet             & ResNet34       & 1.91 & 14.91 & 1.87 & 	14.22 & 1.38 & 5.96 &  1.36 & 	5.90 & 1.19 & 2.26  & \textbf{1.17 }& 	\textbf{1.99}             \\ \midrule
			20 Newsgroups             & Global-Pool CNN       & 602.68 & 14.78 & 559.50 & 	16.53 & 988.42 & \textbf{3.45} &  520.50 & 	17.30 & 729.39 & 13.35  &  \textbf{487.82} & 	16.55  \\ \bottomrule
		\end{tabular}%
	}
	\caption{Our loss is meant to be used in addition to another application specific loss. The table compares the calibration performance improvement using MDCA over three commonly used loss functions (\texttt{NLL/LS/FL}). Our loss improves calibration performance across multiple datasets and architectures.}
	\label{tab:sel-config}
\end{table*}

\begin{table*}[t]
	\centering
	\resizebox{\linewidth}{!}{%
		\begin{tabular}{cc|ccc|ccc|ccc|ccc||ccc}
			\toprule
			&  &  \multicolumn{3}{c|}{\textbf{BS \cite{brierloss}}} & \multicolumn{3}{c|}{\textbf{DCA \cite{dcapaper}}}   & \multicolumn{3}{c|}{\textbf{MMCE \cite{kumarpaper}}} & \multicolumn{3}{c||}{\textbf{FLSD \cite{focallosspaper}}}  & \multicolumn{3}{c}{\textbf{Ours ({FL+MDCA})}} \\
			\multirow{-2}{*}{\textbf{Dataset}} &
			\multirow{-2}{*}{\textbf{Model}} &

			SCE &
			ECE &
			TE &
			SCE &
			ECE &
			TE &
			SCE &
			ECE &
			TE &
			SCE &
			ECE &
			TE &
			SCE &
			ECE & 
			TE \\
			\midrule
			\multirow{2}{*}{CIFAR10}  & ResNet32               &  6.60 & 2.92 & 7.76 &  8.41 & 4.00 & \textbf{7.06} &  8.17 & 3.31 & 8.41 &  9.48 & 4.41 & 7.87 &  \textbf{3.22} & \textbf{0.93} & 7.18 \\
			& ResNet56                &  5.44 & 2.17 & 7.75 &  7.59 & 3.38 & \textbf{6.53} &  9.11 & 3.71 & 8.23 &  7.71 & 3.49 & 7.04 &  \textbf{2.93} & \textbf{0.70} & 7.08                \\ \midrule
			\multirow{2}{*}{CIFAR100} & ResNet32                &  1.97 & 5.32 & 33.53 &  2.82 & 11.31 & 29.67 &  2.79 & 11.09 & 31.62 &  1.77 & 1.69 & 32.15 &  \textbf{1.72} & \textbf{1.49} & \textbf{31.58}         \\
			& ResNet56                &  1.86 & 4.69 & 30.72 &  2.77 & 9.29 & 43.43 &  2.35 & 8.61 & \textbf{28.75} &  1.71 & 1.90 & 29.11 &  \textbf{1.60} & \textbf{0.72} & 29.8     \\ \midrule
			\multirow{2}{*}{SVHN}     & ResNet20                &  2.12 & \textbf{0.45} & \textbf{3.56} &  4.29 & 2.02 & 3.83 &  9.18 & 4.34 & 4.12 &  18.98 & 9.37 & 4.10 &  \textbf{1.90} & 0.47 & 3.92       \\
			& ResNet56              &  2.18 & 0.66 & \textbf{3.25} &  2.16 & 0.49 & 3.32 &  9.69 & 4.48 & 4.26 &  26.15 & 13.23 & 3.65 &  \textbf{1.51} & \textbf{0.23} & 3.85        \\ \midrule
			Mendeley V2               & ResNet50                &  117.6 & 3.75 & 18.43 &  145.1 & 8.29 & 17.47 &  130.4 & \textbf{3.45} & \textbf{15.06} &  104.3 & 9.64 & 19.71 &  \textbf{85.68} & 4.81 & 17.95     \\ \midrule
%			Kather5000                & ResNet34                &  32.76 & 8.64 & 16.50 &  27.13 & 4.49 & 16.80 &  47.35 & 16.63 & 17.90 &  57.87 & 22.03 & 19.30 &  41.39 & 13.45 & 16.60              \\ \midrule
			Tiny-ImageNet             & ResNet34                &  1.53 & 7.79 & 43.00 &  2.11 & 17.40 & \textbf{36.68} &  1.62 & 9.71 & 40.75 &  1.18 & \textbf{1.91} & 37.01 &  \textbf{1.17} & 1.99 & 37.49            \\ \midrule
			20 Newsgroups             & Global-Pool CNN                &  725.82 & 13.71 & \textbf{25.93} &  719.83 & 15.30 & 28.07 &  731.31 & 12.69 & 28.63 &  940.70 & \textbf{4.52} & 30.80 &  \textbf{487.82} & 16.55 & 27.88         \\ \bottomrule
		\end{tabular}%
	}
	\caption{Calibration measures SCE ($10^{-3}$) and ECE (\%) score) and Test Error (TE) (\%) in comparison with various competing methods. We use $M=15$ bins for SCE and ECE calculation. We outperform all the baselines across various popular benchmark datasets, and architectures in terms of calibration, while maintaining a similar accuracy.}
	\label{tab:sce-all-methods}
\end{table*}

\section{Dataset and Evaluation}
\label{sec:exp_setup}

\myfirstpara{Datasets}
We validate our technique on well-known benchmark datasets for image classification, semantic segmentation and natural language processing (NLP). For each of the datasets: CIFAR10/100 \cite{krizhevsky2009learning}, SVHN \cite{netzer2011reading}, Mendeley V2 \cite{kermany2018labeled}, Tiny-ImageNet \cite{imagenetpaper} and 20-Newsgroups \cite{20newsgroup}, we have a separate train and test set. The train set is further split into 2 mutually exclusive sets (a) training set containing $90\%$ of the samples, and (b) the validation set containing $10\%$. We use validation set as the hold-out set for post-hoc calibration. This division has been consistent throughout our experimentation. See Supplementary material for detailed description of datasets, \dnn architectures, and training procedure.

\mypara{Evaluation}
We report calibration measures, \sce, \ece, and class-$j$-\ece along with test error for studying calibration performance. We observe that we achieve near-perfect calibration using our technique without any significant drop in the accuracy. We also visualize the calibration using reliability diagrams (please see supplementary material for detailed description of reliability diagrams).

\mypara{Compared Techniques}
We compare our method against models trained with Cross-Entropy (\nll), Label Smoothing (\ls) \cite{originallabelsmoothing}, \texttt{\dca} \cite{dcapaper}, Focal Loss (\fl) \cite{ogfocalloss}, Brier Score (\bs) \cite{brierloss}, \flsd \cite{focallosspaper} as well as \mmce \cite{kumarpaper}. For details on individual methods and their training specifics, please refer to the supplementary.

\section{Results}
\label{subsec:results}

\myfirstpara{Experiments with Application Specific Loss Functions}
Our loss is meant to be used in conjunction with another  application specific loss function to help improve the calibration performance of a model. Common application specific loss include cross entropy loss (\nll) which in turn minimizes negative log likelihood score of the ground truth label in the predicted confidence vector. Focal Loss (\texttt{FL}) \cite{ogfocalloss} has been proposed to improve training in the presence of many easy negatives, and fewer hard negatives. Whereas Label Smoothing (\ls) \cite{originallabelsmoothing} introduces another term in the \nll to smoothen the prediction of a model. We add the proposed \mdca with each of these loss terms, and measure the calibration performance of a model (in terms of \ece, and \sce scores), before and after adding our loss. \cref{tab:sel-config} shows the result. We refer to configurations using our technique as \textbf{``*+\mdca''}, where * refers to \nll/\ls/\fl. For each of the combination we use relative weight of $\beta \in \{1,5,10,15,20,25\}$ for $\mathcal{L}_\mdca$, and report the calibration performance of the most accurate model on the validation set. Our experiments suggest that setting $\beta < 1$ did not have strong regularizing effect). For $\mathcal{L}_\ls$ we use $\alpha=0.1$, and for $\mathcal{L}_\fl$ we use $\gamma \in \{1,2,3\}$. Please refer to \cite{originallabelsmoothing} and \cite{ogfocalloss} for interpretation of $\alpha$, and $\gamma$ respectively. \cref{tab:sel-config} shows that the proposed \mdca loss improves calibration performance of all the above application specific loss functions, across multiple datasets, and architectures. We also note that \flns+\mdca gives best calibration performance. We use this loss configuration in our experiments hereafter.

\begin{table}[]
	\centering
	\resizebox{0.47 \textwidth}{!}{%
		\begin{tabular}{lcccccccccc}
			\toprule
			\multicolumn{1}{c}{\multirow{2}{*}{\textbf{Method}}} & \multicolumn{10}{c}{\textbf{Classes}}                                                                                                                                             \\  \cmidrule(lr){2-11}
			\multicolumn{1}{c}{}                                 & \textbf{0}      & \textbf{1}      & \textbf{2}      & \textbf{3}      & \textbf{4}      & \textbf{5}      & \textbf{6}      & \textbf{7}      & \textbf{8}      & \textbf{9}      \\ \midrule
			Cross Entropy                                                  & \textbf{0.20}         & 0.62          & 0.33          & 0.65          & 0.23 & 0.36          & 0.25          & 0.26          & \textbf{0.21 }         & 0.41          \\
			Focal Loss \cite{ogfocalloss}                                         & 0.30          & 0.48          & 0.41          & \textbf{0.18} & 0.38          & 0.19          & 0.33          & 0.36          & 0.32          & 0.30          \\
			LS \cite{labelsmoothinghelp}                                                   & 1.63          & 2.60          & 2.54          & 1.90          & 1.91          & 1.74          & 1.73          & 1.75          & 1.63          & 1.58          \\
			Brier Score \cite{brierloss}                                                   & 0.23          & 0.28          & 0.40          & 0.45          & 0.25          & 0.26          & 0.25          & 0.27          & 0.21          & 0.37          \\
			MMCE \cite{kumarpaper}                                                 & 1.78          & 2.35          & 2.12          & 2.00          & 1.74          & 1.87          & 1.65          & 1.76          & 1.70          & 1.84          \\
			DCA \cite{dcapaper}                                                  & 0.31          & 0.70          & 0.40          & 0.72          & 0.31          & 0.46          & 0.35          & 0.35          & 0.37          & 0.36 
			\\
			FLSD \cite{focallosspaper}  & 1.52 & 3.24 & 2.74 & 2.15 & 1.79 & 1.82 & 1.84 & 1.62 & 1.54 & 1.38 \\ \midrule
			% NLL+MDCA (ours)                                           & \textbf{0.08} & \textbf{0.18} & 0.18 & 0.23          & 0.98          & \textbf{0.18} & \textbf{0.12} & \textbf{0.16} & \textbf{0.14} & \textbf{0.23} \\
			% LS+MDCA (ours)                                              & 1.13          & 2.09          & 1.73          & 1.35          & 1.42          & 1.20          & 1.18          & 1.39          & 1.17          & 1.11         \\
			\rowcolor{LightCyan}
			\textbf{Ours (FL+MDCA)}          & 0.22 & \textbf{0.16} & \textbf{0.24} & 0.25 & \textbf{0.22} & \textbf{0.16} & \textbf{0.16} & \textbf{0.17}   & 0.25 & \textbf{0.20} \\
			\bottomrule
		\end{tabular}%
	}
	\caption{Class-$j$-ECE (\%) score on all $10$ classes for ResNet-20 model trained on the SVHN dataset with different learnable calibration methods. Our method gives best calibration for 7 out of 10 classes, and is second-best on 3 classes.}
	\label{tab:classwiseECE}
\end{table}

%\begin{figure}[t]
%	\centering
%	\includegraphics[width =\linewidth]{figs/cifar10_overconfident.pdf}
%	%\vspace{-4 mm}
%	\caption{Reliability diagrams (a,b) and confidence histograms (c,d) of \nll trained model compared against MDCA regularized version (\texttt{NLL+MDCA}). We use ResNet-32 trained on CIFAR10 dataset for comparison. Please refer to the text for the interpretation of the plots.}
%	\label{fig:relDiagramNLL}
%\end{figure}
%
%\begin{figure}[t]
%	\centering
%	\includegraphics[width =\linewidth]{figs/svhn_underconfident.pdf}
%	%\vspace{-4 mm}
%	\caption{Reliability diagrams (a,b) and confidence histograms (c,d) of ResNet-20 network trained with Label Smoothing(\ls) vs. \mdca regularized \ls on SVHN dataset.}
%	\label{fig:relDiagramLS}
%\end{figure}

\begin{figure}[t]
	\centering
	\def\imgwidth{0.23\linewidth}
	\begin{subfigure}{\imgwidth}
		\centering
		\includegraphics[width=\linewidth]{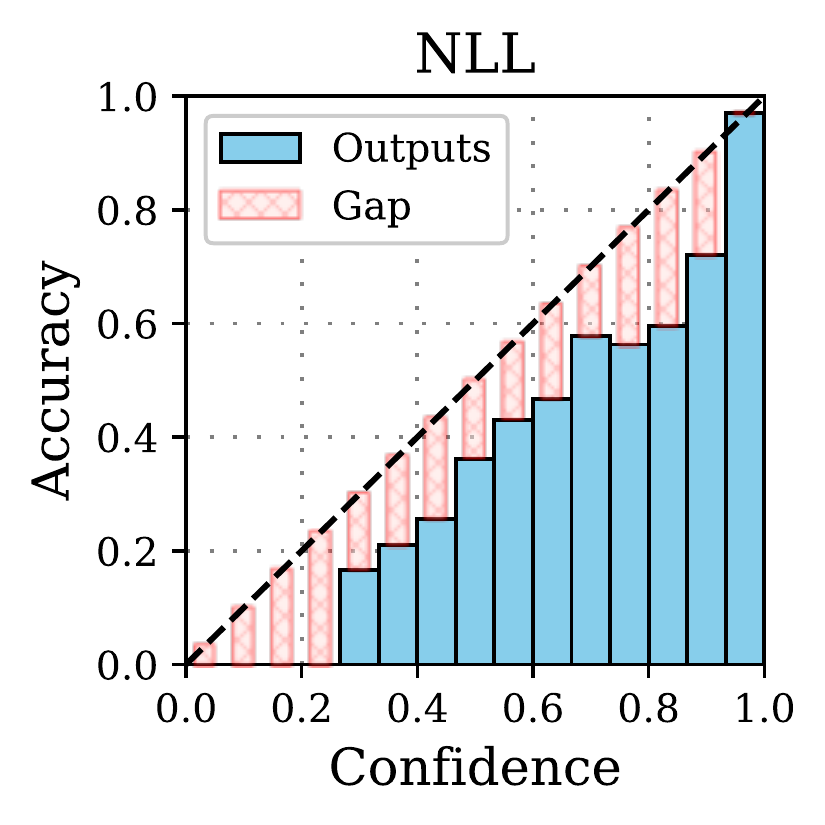}
		\caption{}
		\label{fig:rel_conf_plots_a}
	\end{subfigure}
	\begin{subfigure}{\imgwidth}
		\centering
		\includegraphics[width=\linewidth]{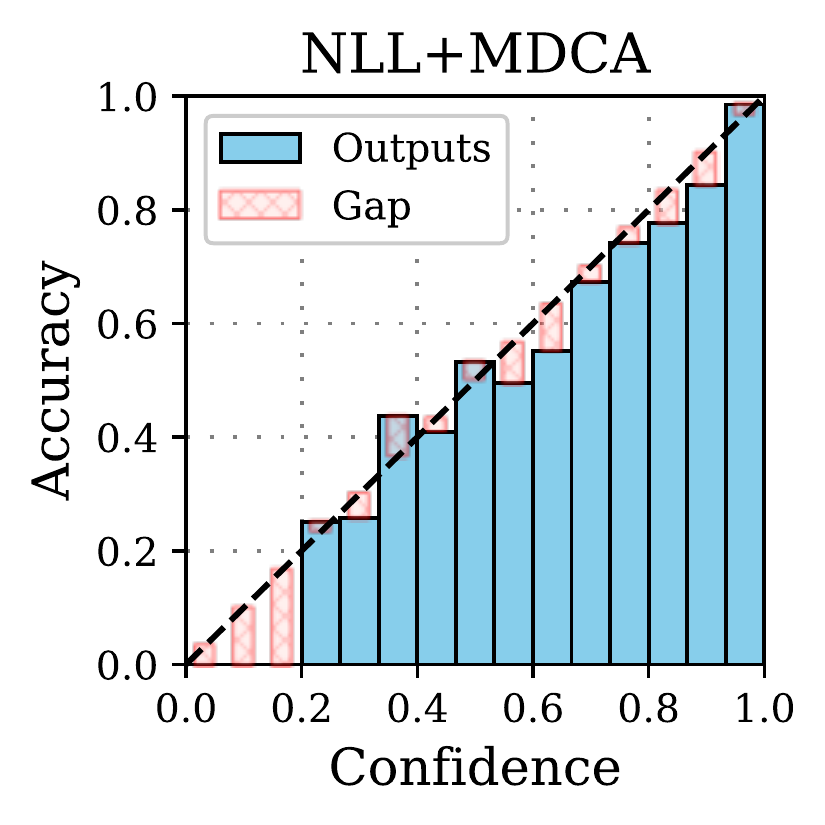}
		\caption{}
		\label{fig:rel_conf_plots_b}
	\end{subfigure}
	\begin{subfigure}{\imgwidth}
		\centering
		\includegraphics[width=\linewidth]{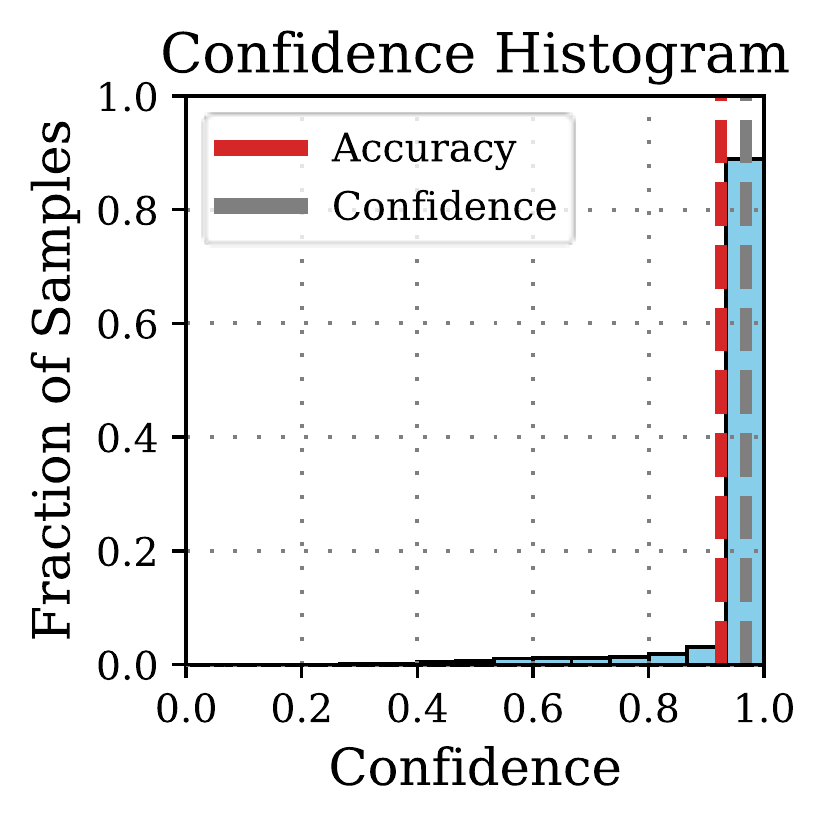}
		\caption{}
		\label{fig:rel_conf_plots_c}
	\end{subfigure}
	\begin{subfigure}{\imgwidth}
		\centering
		\includegraphics[width=\linewidth]{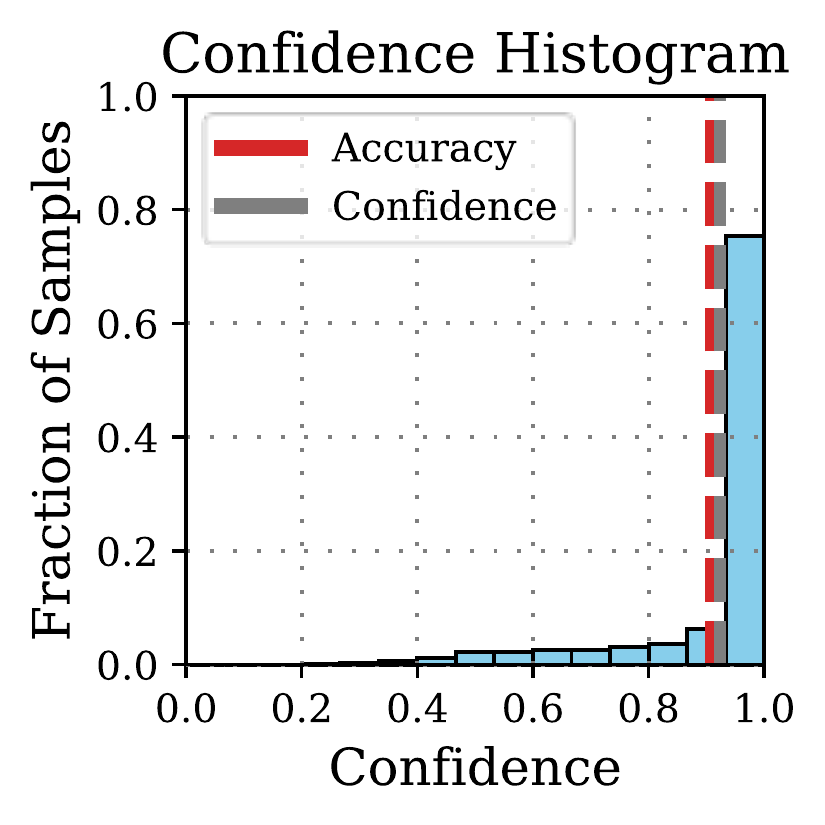}
		\caption{}
		\label{fig:rel_conf_plots_d}
	\end{subfigure}
	\begin{subfigure}{\imgwidth}
		\centering
		\includegraphics[width=\linewidth]{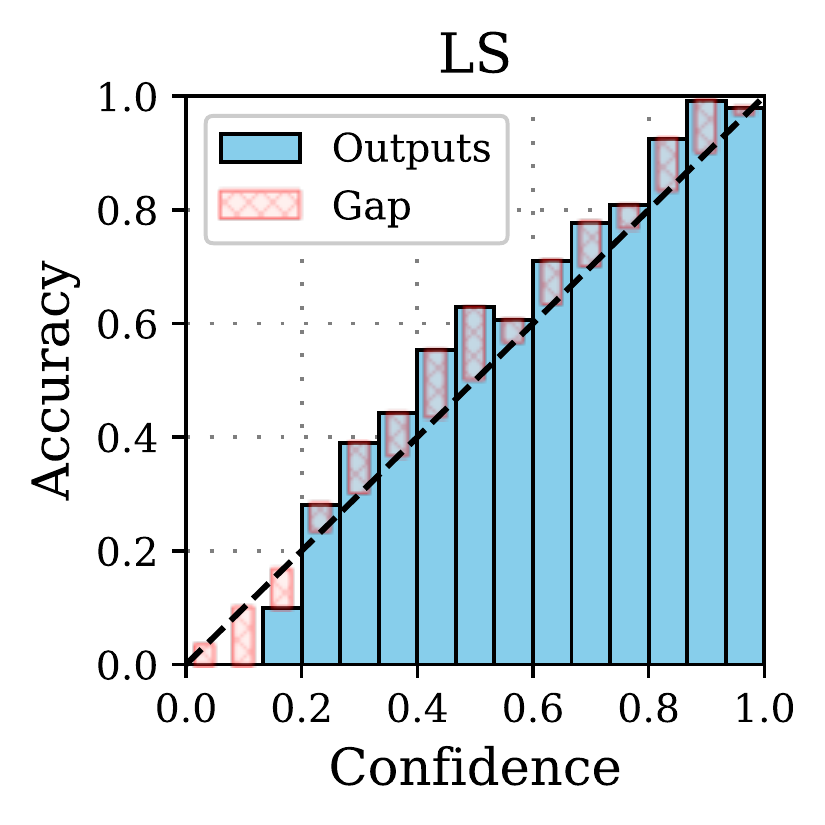}
		\caption{}
		\label{fig:rel_conf_plots_e}
	\end{subfigure}
	\begin{subfigure}{\imgwidth}
		\centering
		\includegraphics[width=\linewidth]{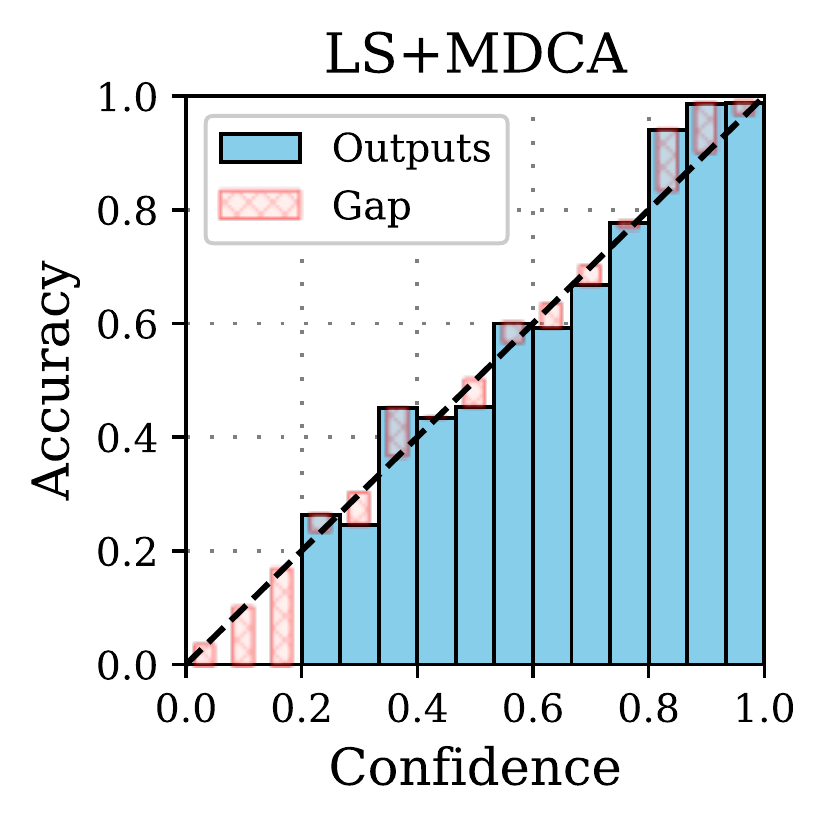}
		\caption{}
		\label{fig:rel_conf_plots_f}
	\end{subfigure}
	\begin{subfigure}{\imgwidth}
		\centering
		\includegraphics[width=\linewidth]{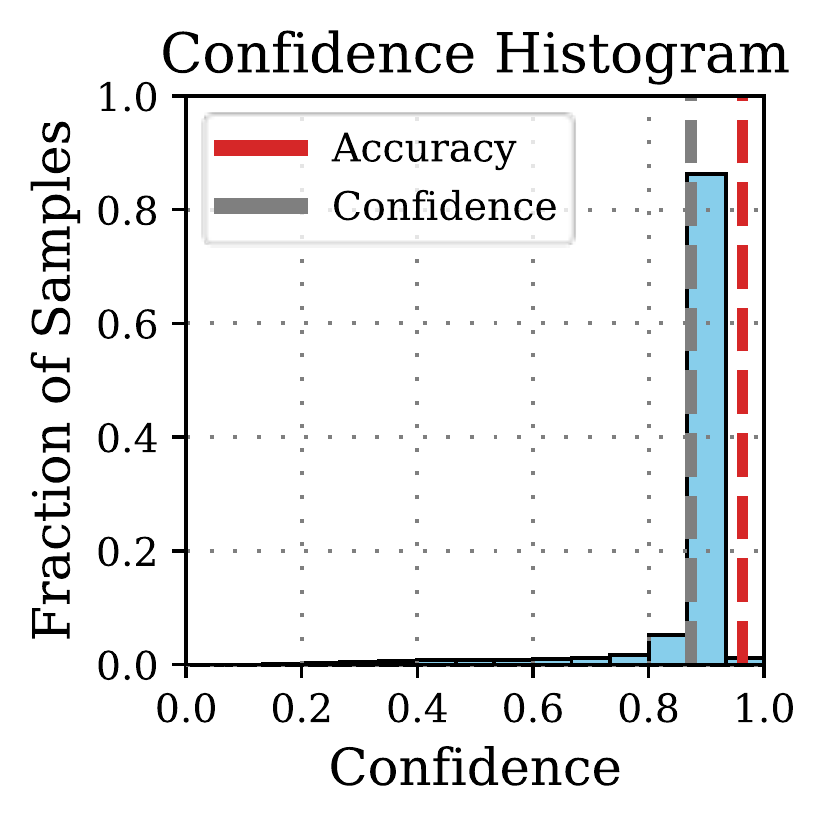}
		\caption{}
		\label{fig:rel_conf_plots_g}
	\end{subfigure}
	\begin{subfigure}{\imgwidth}
		\centering
		\includegraphics[width=\linewidth]{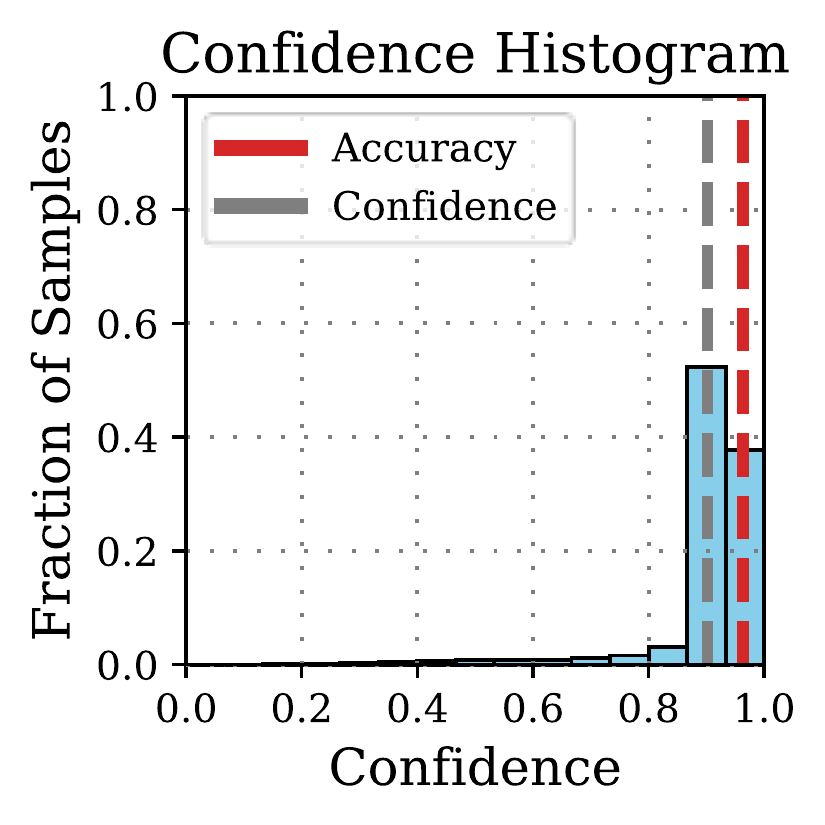}
		\caption{}
		\label{fig:rel_conf_plots_h}
	\end{subfigure}	
	\caption{First row shows Reliability diagrams (a,b) and confidence histograms (c,d) of \nll trained model compared against \mdca regularized version (\nllns+\mdca). We use ResNet-32 trained on CIFAR10 dataset for comparison. Second row shows corresponding plots for ResNet-20 network trained with Label Smoothing(\ls) vs. \mdca regularized \ls on SVHN dataset. Please refer to the text for the interpretation of the plots. We show a similar comparison with \fl, and \flns+\dca in the supplementary.}
	\label{fig:rel_conf_plots}
\end{figure}

% \begin{figure}[t]
% 	\centering
% 	\includegraphics[width =\linewidth]{figs/Compact Reliability and confidence histograms.pdf}
% 	%\vspace{-2 mm}
% 	\caption{First row shows Reliability diagrams (a,b) and confidence histograms (c,d) of \nll trained model compared against \mdca regularized version (\nll+\mdca). We use ResNet-32 trained on CIFAR10 dataset for comparison. Second row shows reliability diagrams (e,f) and confidence histograms (g,h) of ResNet-20 network trained with Label Smoothing(\ls) vs. \mdca regularized \ls on SVHN dataset. Please refer to the text for the interpretation of the plots. \textbf{Pls. refer to suppl. material where we demonstrate the superiority of \texttt{FL+MDCA} over \texttt{FL+DCA} as well as \texttt{FL} only.}}
% 	\label{fig:compact_rel_diags}
% \end{figure}

\mypara{Calibration Comparison with \sota}
\cref{tab:sce-all-methods} compares calibration performance of our method with all recent \sota methods. We note that calibration using our method improves both \sce as well as \ece score on all the datasets, and different architectures. 

\mypara{Class-Conditioned Calibration Error}
Current state-of-the-art focuses on calibrating the predicted label only, which leaves some of the minority class un-calibrated. One of the benefits of our calibration approach is better calibration for all and not only the predicted class. To demonstrate the effectiveness of our method, we report class-$j$-\ece \% values of all the competing methods against our method, using ResNet-20 model trained on the \svhn dataset. \cref{tab:classwiseECE} shows the result. Our method gives best scores for all but 3 out of 10 classes, where it is second-best. Class-wise reliability diagrams (c.f. \cref{fig:teaser}) reinforce a similar conclusion. We show results on \cifar10 dataset in the supplementary.

\mypara{Test Error} 
\cref{tab:sce-all-methods} also shows the Test Error (\te) obtained by a model trained using our method and other \sota approaches. We note that using our proposed loss, a model is able to achieve best calibration performance without sacrificing on the prediction accuracy (Test Error). %\textcolor{red}{ Although it might be noted that we are not significantly improving on the test accuracy, we do improve the calibration for the predictions in the hope that a promising post processing step will be able to apply corrective changes to the predicted probabilities in order to give more accurate results. This might include post-processing via a Conditional Random Field, etc. However, in this paper we abstain from exploring the post-processing and focus on the calibration. Moreover, it can be observed that we are in a comparable range with other commonly used methods.}  % Our methods obtain comparable test error \% to other methods.

\myfirstpara{Mitigating Under/Over-Confidence}
\cref{tab:sel-config} and \cref{tab:sce-all-methods} already show that our method improves over \sota in terms of \sce, and \ece scores. However the tables do not highlight whether they correct for over-confidence or under-confidence. We show the reliability diagram (\cref{fig:rel_conf_plots}) for a ResNet-32 model trained on \cifar10. The uncalibrated model is overconfident (\cref{fig:rel_conf_plots_a}) which gets rectified after calibrating with our method (\cref{fig:rel_conf_plots_b}). We also show confidence plots in the picture, and the colored dashed lines to indicate average confidence of the predicted label, and the accuracy. It can be seen that accuracy is lower than average confidence in the uncalibrated confidence plot (\cref{fig:rel_conf_plots_c}), which indicates the overconfident model. After calibrating with our method, the two dashed lines almost overlap indicating the perfect calibration achieved (\cref{fig:rel_conf_plots_d}). Similarly, second row of \cref{fig:rel_conf_plots} show that the model trained with \ls solely is under-confident; and a model trained with \ls along with \mdca is confident and calibrated.

%\begin{SCfigure}
\begin{figure}[t]
	\centering
	\vspace{-1em}
	\includegraphics[width=0.8\linewidth]{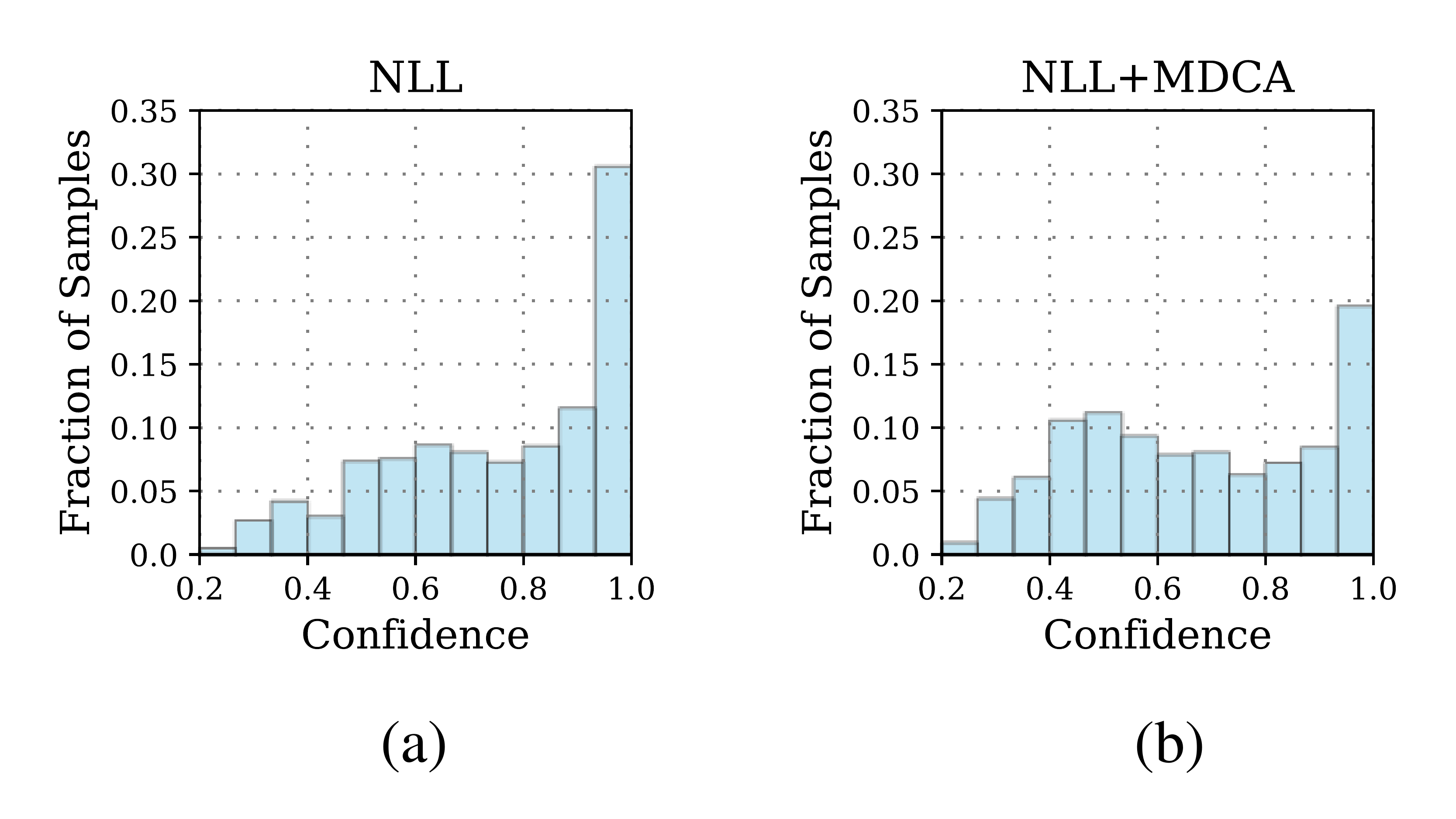}
	\vspace{-1em}
	\caption{Confidence value histogram for misclassified predictions. \mdca regularized \nll makes less confident incorrect predictions as compared to the uncalibrated method trained using \nll.}
	\label{fig:missclassified-conf-hist}
\end{figure}

\mypara{Confidence Values for Incorrect Predictions} 
The focus of the discussion so far has been on the fact that confidence value for a class should be consistent with the likelihood of the class for the sample. Here, we analyze our method for the confidence values it gives when the prediction is incorrect. \cref{fig:missclassified-conf-hist} shows the confidence value histogram for all the incorrect predictions made by the ResNet-32 model trained on \cifar10 dataset using \nll vs. \mdca regularized \nll. It is clear that our calibration reduces the confidence for the mis-prediction. The same is also evident from the \cref{fig:teaser} shown earlier.

\begin{table}
	\centering
	\resizebox{1.0\linewidth}{!}{%
	\setlength{\tabcolsep}{15pt}
		\begin{tabular}{lccc|c}
			\toprule
			\multicolumn{1}{c}{\textbf{Method}} & \textbf{Art}    & \textbf{Cartoon} & \textbf{Sketch}  & \textbf{Average} \\ \midrule
			NLL                                 & 6.33          & 17.95          & 15.01          & 13.10              \\
			LS \cite{labelsmoothinghelp}                                  & 7.80          & 11.95          & \textbf{10.88} & 10.21              \\
			FL \cite{ogfocalloss}                                  & 8.61          & 16.62          & 10.94          & 12.06              \\
			Brier Score \cite{brierloss}                               & 6.55          & 13.19          & 15.63          & 11.79              \\
			MMCE \cite{kumarpaper}                                & 6.35          & 15.70          & 17.16          & 13.07              \\
			DCA \cite{dcapaper}                                 & 7.49          & 18.01          & 14.99          & 13.49              \\ 
			FLSD \cite{focallosspaper}                       & 8.35          & 13.39          & 13.86          & 11.87              \\
			\midrule
			%\rowcolor{LightCyan}
			%NLL+MDCA                           & \textbf{5.10} & 16.53          & 13.49          & 11.71              \\
			%\rowcolor{LightCyan}
			%LS+MDCA                            & 5.70          & 12.07          & 11.70          & 9.82               \\
			\rowcolor{LightCyan}
			Ours (FL+MDCA)                          & \textbf{6.21}          & \textbf{11.91} & 11.08          & \textbf{9.73} \\
			\bottomrule
		\end{tabular}%
	}
	\caption{Calibration performance (\sce($10^{-3}$)) under domain shift on \pacs dataset \cite{pacspaper}. For each column we train on the other two subsets, and then test on the subset listed as the column heading.}
	\label{tab:datasetdrift}
\end{table}
\begin{table}[t]
	\centering
	\resizebox{1.0\linewidth}{!}{%
	\setlength{\tabcolsep}{15pt}
	\begin{tabular}{l|ccc|c}
		\toprule
		\multicolumn{1}{c|}{\multirow{2}{*}{\textbf{Method}}} & \multicolumn{3}{c|}{\textbf{CIFAR10}}    & \textbf{SVHN} \\
		& \textbf{IF-10}  & \textbf{IF-50}          & \textbf{IF-100}  &         \textbf{IF-2.7}            \\ \midrule
		{NLL}                        & 18.44           & 32.21  & 31.04               & 3.43                  \\
		{FL \cite{ogfocalloss}}                         & 14.65           & 29.67  & 28.89               & 2.54                  \\
		{LS \cite{labelsmoothinghelp}}                         & 14.88           & 26.30  & \textbf{20.79}      & 18.80                 \\
		{BS \cite{brierloss}}                         & 15.74           & 33.57  & 29.01               & 2.12                  \\
		{MMCE \cite{kumarpaper}}                       & 15.10           & 29.05  & 21.56               & 9.18                  \\
		{FLSD \cite{focallosspaper}}                       & 16.05           & 31.35  & 30.28               & 18.98                 \\
		{DCA \cite{dcapaper}}                        & 18.57           & 32.81  & 35.53               & 4.29                  \\ \midrule
		\rowcolor{LightCyan}
		Ours (\texttt{FL+MDCA})                 & \textbf{11.83}    & \textbf{22.97} & 26.89       & \textbf{1.90} \\
		
		\bottomrule
	\end{tabular}%
	}
	\caption{Our calibration technique works best even when there is a significant class imbalance in the dataset. For this experiment we created imbalance of various degrees in \cifar10 as suggested in \cite{skew}. Original \svhn has a Imbalance Factor(IF) of $2.7$. Hence we show calibration performance (\sce($10^{-3}$)) on original \svhn.}
	\label{tab:sce-data-imbalance}
\end{table}

\mypara{Calibration Performance under Dataset Drift} 
Tomani \etal \cite{domain-drift-calibration-posthoc} show that \dnns are over-confident and highly uncalibrated under dataset/domain shift. Our experiments shows that a model trained with \mdca fairs well in terms of calibration performance even under non-semantic/natural domain shift. We use two datasets (a) \pacs \cite{pacspaper} and (b) Rotated \mnist inspired from \cite{can-u-trust}. The datasets are benchmarks for synthetic non-semantic shift and natural rotations respectively. Dataset specifics and training procedure are provided in the supplementary. \cref{tab:datasetdrift} shows that our method achieves the best average \sce value across all the domains in \pacs. A similar trend is observed on Rotated \mnist dataset as well (see supplementary), where our method achieves the least average \sce value across all rotation angles. 

\mypara{Calibration Performance on Imbalanced Datasets}
The real-world datasets are often skewed and exhibit long-tail distributions, where a few classes dominate over the rare classes. In order to study the effect of class imbalance on the calibration quality, we conduct the following experiment, where we introduce a deliberate imbalance on \cifar10 dataset to force a long-tail distribution as detailed in \cite{skew}. \cref{tab:sce-data-imbalance} shows that a model trained with our method has best calibration performance in terms of \sce score across all imbalance factors. We observe that \svhn dataset already has a imbalance factor of $2.7$, and hence create no artificial imbalance in the dataset for this experiment.
The efficacy of our approach on the imbalanced data is
%on two accounts: first, with the implicit regularisation offered by FL (acts as a maximum-entropy regularizer) and second, {importantly 
due to the regularization provided by \mdca which penalizes the difference between average confidence and average count even for the non-predicted class, hence benefiting minority classes.

\begin{table}[t]
	\centering
	\resizebox{1.0\linewidth}{!}{%
		\setlength{\tabcolsep}{14pt}
	\begin{tabular}{lcccc}
		\toprule
		\multirow{2}{*}{\textbf{Method}} & \multirow{2}{*}{\textbf{Post Hoc}} 
		& & \textbf{SCE($10^{-3}$)} $\downarrow$ & \\
		\cline{3-5}
		& & \textbf{CIFAR10}  & \textbf{CIFAR100} & \textbf{SVHN} \\ \midrule
		\multirow{3}{*}{NLL}   & None & 7.12  & 2.50 & 3.84  \\
		& TS   & 3.25  & 1.49 & 4.16  \\
		& DC   & 4.98  & 1.91 & 2.69  \\ \midrule
		\multirow{3}{*}{LS \cite{labelsmoothinghelp}}    & None & 12.55 & 1.73 & 21.08 \\
		& TS   & 4.49  & 1.67 & 3.12  \\
		& DC   & 5.34  & 1.98 & 2.81  \\ \midrule
		\multirow{3}{*}{FL \cite{ogfocalloss}}    & None & 4.19  & 1.89 & 7.85  \\
		& TS   & 4.19  & 1.62 & 2.72  \\
		& DC    & 5.48  & 2.02 & 3.36  \\ \midrule
		\multirow{3}{*}{BS \cite{brierloss}} & None & 5.44  & 1.86 & 2.18  \\
		& TS   & 3.94  & 1.68 & 3.88  \\
		& DC   & 4.83  & 1.80 & 2.11  \\ \midrule
		\multirow{3}{*}{MMCE \cite{kumarpaper}}  & None & 9.12  & 2.35 & 9.69  \\
		& TS   & 4.05  & 1.61 & 3.74  \\
		& DC    & 6.26  & 1.95 & 5.11  \\ \midrule
		\multirow{3}{*}{DCA \cite{dcapaper}}   & None & 7.60  & 2.87 & 2.16  \\
		& TS   & 3.00  & 1.56 & 4.29  \\
		& DC    & 4.20  & 2.06 & 2.95  \\ \midrule
		\multirow{3}{*}{FLSD \cite{focallosspaper}}  & None & 7.71  & 1.71 & 26.15 \\
		& TS   & 3.27  & 1.71 & 4.41  \\
		& DC   & 5.62  & 2.01 & 4.31  \\ \midrule
		\multirow{3}{*}{Ours (FL+MDCA)}         & \cellcolor[HTML]{E0FFFF}None                               & \cellcolor[HTML]{E0FFFF}\textbf{2.93}     & \cellcolor[HTML]{E0FFFF}\textbf{1.60}     & \cellcolor[HTML]{E0FFFF}\textbf{1.51}     \\
		& TS   & 2.93  & 1.60 & 5.00  \\
		& DC   & 3.81  & 1.87 & 2.72 \\ \bottomrule
	\end{tabular}%
	}
	\caption{Results after combining various trainable calibration methods including ours with two post-hoc calibration methods (\ts: Temperate scaling \cite{platt1999probabilistic}, and \dc: Dirichlet Calibration \cite{Dirichlet}) on \sce($10^{-3}$). We use ResNet56 model on \cifar10, \cifar100, and \svhn datasets for this experiment. Though other methods benefit by post-hoc calibration, our method outperforms them without using any post-hoc calibration.}
	\label{tab:TrainVsPH}
\end{table}

\mypara{Our Approach + Post-hoc Calibration} 
We study the performance of combined effect of post-hoc calibration methods, namely Temperature Scaling (\ts) \cite{platt1999probabilistic}, and Dirichlet Calibration (\dc) \cite{Dirichlet}, applied over various train-time calibration methods including ours (\flns+\mdca). \cref{tab:TrainVsPH} shows the results. We observe that while \ts, and \dc improve the performance of other competitive methods, our method outperforms them even without using any of these methods. On the other hand, the performance of our method seems to either remain same or slightly decrease after application of post-hoc methods. We speculate that this is because our method already calibrates the model to near perfection. For example, on performing \ts, we observe the optimal temperature values are $T \approx 1$ implying that it leaves little scope for the TS to improve on top of it. Thus, any further attempt to over-spread the confidence prediction using \ts or \dc negatively affects the confidence quality.

\begin{table}[t]
	\resizebox{\linewidth}{!}{%
		\begin{tabular}{lcccc}
			\toprule
			\multicolumn{1}{c}{\textbf{Method}}         & \textbf{Pixel Acc. (\%)} & \textbf{mIoU (\%)} & \textbf{SCE ($10^{-3})$}    & \textbf{ECE (\%) }    \\ 
			\midrule
			\nll                      & 94.81              & 79.49        & 6.4          & 7.77          \\
			\nllns+\ts        & 94.81              & 79.49        & 6.26          & 6.1          \\
			% NLL+MDCA          & 94.71             & 79.39        & \textbf{6.3} & \textbf{7.71}          \\ 
			\fl             & 92.85              & 77.22        & 11.8          & 7.69          \\
			% FL+MDCA ($\gamma = 3$, $\beta =  1$) & 93.03              & 77.45        & 9.3          & 4.91          \\
			\textbf{Ours (\flns+\mdca)}  & 94.47              & 78.66        & \textbf{5.8} & \textbf{4.66}  \\ 
			\bottomrule
		\end{tabular}%
	}
	\caption{Segmentation results on Xception65 \cite{xception} backbone DeeplabV3+ model \cite{deeplabv3Plus} on \pascal 2012 validation dataset.}
	\label{tab:segmentation-table}
\end{table}

\myfirstpara{Calibration Results for Semantic Segmentation} 
One of the major advantages of our technique is that it allows to use billions of weights of a \dnn model to be used for the calibration. This is in contrast to other calibration approaches which are severely constrained in terms of parameters available for tuning. For example in \ts one has a single temperature parameter to tune. This makes it hard for \ts to provide image and pixel specific confidence transformation for calibration. To highlight pixel specific calibration aspect of our technique we have done experiments on semantic segmentation task, which can be seen as pixel level classification. For the experiment, we train a DeepLabV3+ \cite{deeplabv3Plus} model with a pre-trained Xception65 \cite{xception} backbone on the \pascal 2012 \cite{pascal-voc-2012} dataset. We compare the performance of our method against \nll, \fl and \ts (post-hoc calibration). Please refer to the supplementary for more details on the training. \cref{tab:segmentation-table} shows the results. We see a significant drop in both \sce/\ece in case of our method (\flns+\mdca) as compared to \fl ($2\times$ drop in \sce and a 40\% decrease in \ece). Our method also outperforms \ts (after training with \nll) by $23.6\%$. 
%When performing \ts post-processing after training with our method, we get a temperature of 1, which confirms our speculation in the previous section (our approach+post-hoc calibration) that \mdca trained model is already well calibrated to begin with and does not require any more post-hoc calibration.

\section{Ablation Study}

\begin{figure}[t]
	\begin{center}
		\includegraphics[width =\linewidth]{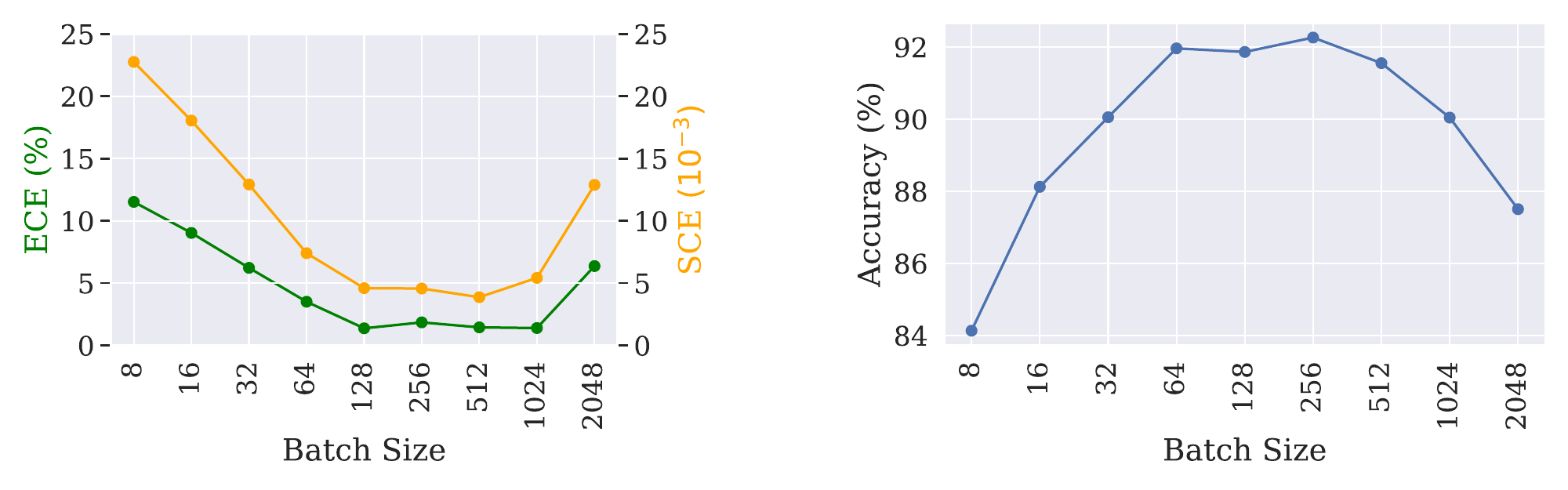}
		\caption{Effect of different batch sizes on Calibration performance metrics (\ece/\sce/Accuracy) while training with \mdca on a ResNet-32 model on \cifar10 dataset. The calibration performance drops with larger batch size because SGD optimization is more effective in a small-batch regime\cite{Keskar2016}. A larger batch results in a degradation in the quality of the model, as measured by its ability to generalize. The performance degradation is also consistent with the model trained using solely on \fl on a large batch size.}
		\label{fig:abl_batchsize}
	\end{center}%
\end{figure}

\mypara{Effect of Batch Size}
\cref{fig:abl_batchsize} shows the effect of different batch sizes on the calibration performance. We vary the batch sizes exponentially and observe that a model trained with \mdca achieves best calibration performance around batch size of 64 or 128. As we decrease (or increase) the batch size, we see a degradation in calibration, though the drop is not significant. 

\begin{figure}[t]
	\begin{center}
	\includegraphics[width =\linewidth ]{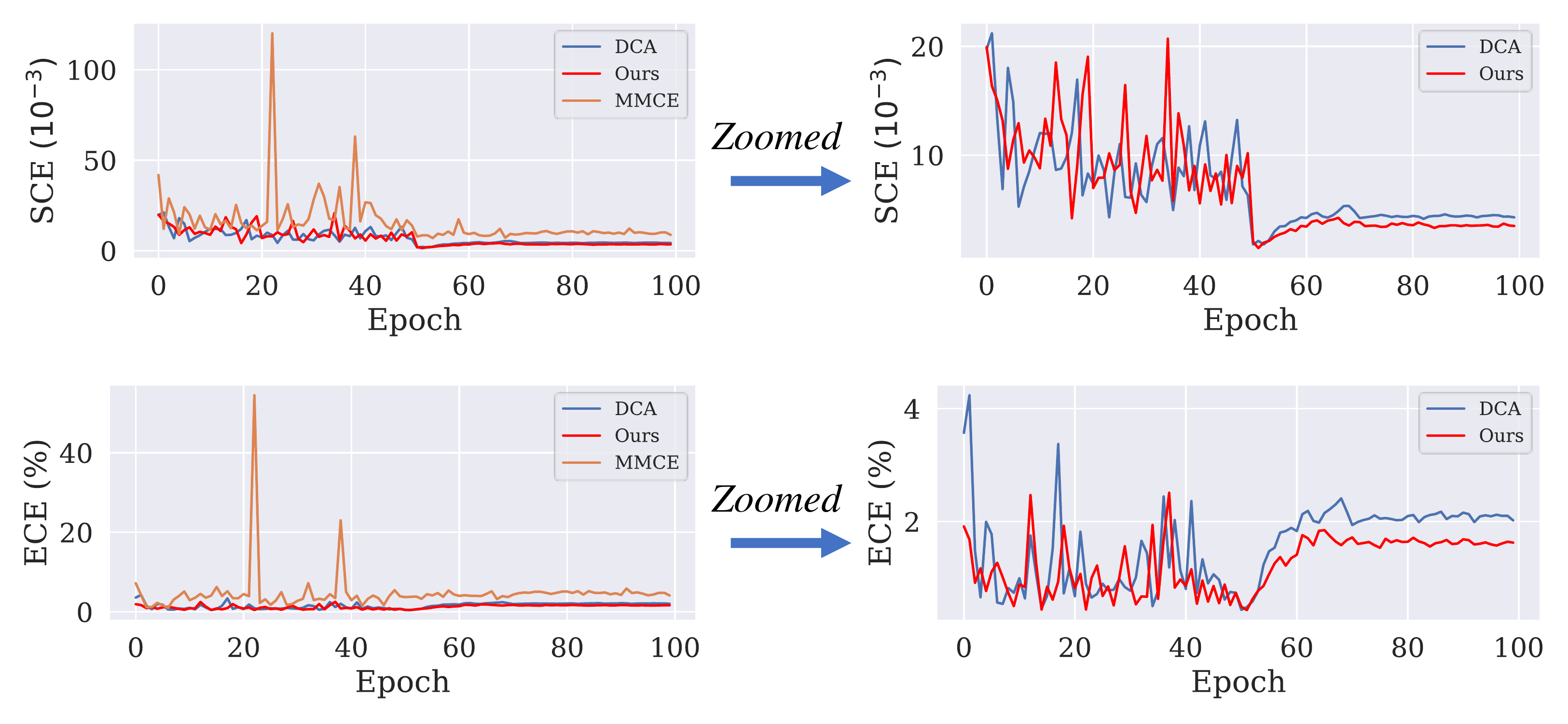}
	\caption{Comparison of \ece/\sce at various epochs for \mdca, \mmce, and \dca. Though, \mmce, and \dca directly optimize for \ece, their loss function is not differentiable and hence the techniques are not able to reduce \ece as much as \mdca. Differentiability of loss function allows \mdca to reduce \ece better even when it does not directly optimize it. We use a learning rate decay of $1/10$ at epochs $50$ and $70$. Please refer to the supplementary for the details of the experiment.}
	\label{fig:abl_ece_convergence}
	\end{center}%
\end{figure}
	
\mypara{Comparison of \ece/\sce Convergence with \sota}
In previous sections, we compared the \ece scores of \mdca with other contemporary trainable calibration methods like \mmce \cite{kumarpaper} and \dca \cite{dcapaper}. Many of these methods explicitly aim to reduce \ece scores. While \mdca does not directly optimize \ece, yet we see in our experiments that \mdca manages to get better \ece scores at convergence. We speculate that this is due to the differentiablity of \mdca loss which helps optimize the loss better using backpropagation. To verify the hypothesis, we plot the \ece convergence for various methods in \cref{fig:abl_ece_convergence}.

\section{Conclusion \& Future work}
\label{concAndLimitations}

We have presented a train-time technique for calibrating the predicted confidence values of a \dnn based classifier. Our approach combines standard classification loss functions with our novel auxiliary loss named, Multi-class Difference of Confidence and Accuracy (\mdca). Our proposed loss function when combined with focal loss yields the least calibration error among both trainable and post-hoc calibration methods. We show promising results in case of long tail datasets, natural/synthetic dataset drift, semantic segmentation and a natural language classification benchmark too. In future we would like to investigate the role of class hierarchies to develop cost-sensitive calibration techniques.

%\section{Acknowledgments}
%Thanks to Harshad Khadilkar and Mayank Baranwal for suggestions.

% \textcolor{red}{Improve below ?}
% \texttt{MDCA} aims to improve SCE. Since SCE subsumes top label calibration it results in better convergence of ECE. We will look into proofs of equivalence in our future work.
% Furthermore for \texttt{MDCA}, the gradients are back-propagated in a sample/batch specific manner.
%\texttt{MDCA} is indeed a differentiable version of SCE as described in the paper.
% 

%{\small
%\bibliographystyle{ieee_fullname}
%\bibliography{egbib}
%}

\appendix

\clearpage

\title{{\huge Supplementary Material:}}
{\large A Stitch in Time Saves Nine: A Train-Time Regularizing Loss for Improved Neural Network Calibration}

\maketitle

\section*{Table of Contents}
\vspace{2em}

\noindent Due to the restrictions on the length, we could not include descriptions of the state-of-the-art techniques, datasets, hyper-parameters and additional results in the main manuscript. However, to keep the overall manuscript self-contained, we include the following in the supplementary material: 
\begin{itemize}
	\item Section 1: Description of competing methods.
	\item Section 2: Detailed description of benchmark datasets used in our experiments.
	\item Section 3: Training procedure and implementation of proposed, and competing methods.
	\item Section 4: Additional results.
	\item Source code link: \href{https://github.com/mdca-loss/MDCA-Calibration}{MDCA Official PyTorch Github}
\end{itemize}

%Note that the supplementary material has been created as an extension of the main manuscript, and hence the Figure, and Table numbers resume from the main-manuscript.

\section{Competing Methods and hyperparameters used}
\label{sec:compMethods}

In this section, we provide a brief description of each of the compared method with hyperparameter settings used in training.

\begin{itemize}
	\item For \texttt{Brier Score} \cite{brierloss} we  train on the loss defined as the squared loss of the predicted probability vector and the one-hot target vector.
	\item \texttt{Label Smoothing} \cite{labelsmoothinghelp}, takes the form $\texttt{LS} = - \sum_{i=1}^N \sum_{j=1}^K q_{i,j} \log (\hat{p}_{i,j})$ where $\hat{p}_{i,j}$ is the predicted confidence score for sample $i$, for class $j$. Similarly, we define soft target vector, \textbf{${q_{i}}$}, for each sample $i$, such that $q_{i,j}=\alpha/(K-1)$ if $j \neq y_i$, else $q_{i,j} = (1-\alpha)$. Here $\alpha$ is a hyper-parameter. We trained using $\alpha=0.1$, and refer to label smoothing as \texttt{LS} in the results.
	\item \texttt{Focal loss} \cite{ogfocalloss} is defined as $\texttt{FL} = - \sum_{i=1}^N (1-\hat{p}_{i,y_i})^{\gamma} \log (\hat{p}_{i,y_i})$, where $\gamma$ is a hyper-parameter. We trained using $\gamma \in \{1,2,3\}$, and report it as \texttt{FL} in the results.
	\item For DCA \cite{dcapaper}, we train on the following loss: $\texttt{NLL} + \beta \cdot \texttt{DCA}$, where \texttt{DCA} is as defined in \cite{dcapaper}, and $\beta$ is a hyper-parameter. We train varying $\beta  \in  \{1,5,10,15,20,25\}$ as performed in \cite{dcapaper}. DCA results are reported under the name \texttt{DCA}. 
	\item We use \texttt{MMCE} \cite{kumarpaper}, as a regularizer along with \texttt{NLL}. We use the weighted \texttt{MMCE} loss in our experiments with $\lambda \in \{2,4\}$.
	\item For \texttt{FLSD} \cite{focallosspaper}, we train with $\gamma=3$. 
\end{itemize}%I think they are done
For each of the above methods, we report the result of the best performing trained model according to the accuracy obtained on the validation set.

\section{Dataset description}
\label{sec:dataset-desc}

We have used the following datasets in our experiments:

\begin{enumerate}
	\item \textbf{CIFAR10} \cite{krizhevsky2009learning}: This dataset has $60,000$ color images of size $32 \times 32$ each, equally divided into $10$ classes. The pre-divided train set comes with $50,000$ images and the test set has around $10,000$ images. Using the policy defined above, we have a train/val/test split having $45000/5000/10000$ images respectively.
	\item \textbf{CIFAR100} \cite{krizhevsky2009learning}: This dataset comprises of $60,000$ colour images of size $32x32$ each, but this time, equally divided into $100$ classes. The pre-divided train set again comes with $50,000$ images and the test set has around $10,000$ images. We have a train/val/test split having $45000/5000/10000$ images respectively.
	\item \textbf{SVHN} \cite{netzer2011reading} : The Street View House Number (SVHN) is a digit classification benchmark dataset that contains $600000$ $32\times 32$ RGB images of printed digits (from $0$ to $9$) cropped from pictures of house number plates. The cropped images are centered in the digit of interest, but nearby digits and other distractors are kept in the image. SVHN has comes with a training set ($73257$) and a testing set ($26032$). We randomly sample $10\%$ of the training set to use as a validation set.
	\item \textbf{Tiny-ImageNet} \cite{imagenetpaper} : It is a subset of the ImageNet dataset containing $64 \times 64$ RGB images. It has $200$ classes with each class having $500$ images. The validation set contains $50$ images per class. We use the provided validation set as the test set for our experiments. 
	\item \textbf{20 Newsgroups} \cite{20newsgroup}: It is a popular text classification dataset containing $20,000$ news articles, categorised evenly into $20$ different newsgroups based on their content.
	\item \textbf{Mendeley V2} ~\cite{kermany2018labeled}: Inspired from \cite{dcapaper}, we use this medical dataset. The dataset contains OCT (optical coherence tomography) images of retina and pediatric chest X-ray images. However, we only use the chest X-ray images in our experiments. The chest X-ray images come with a pre-defined train/test split having $4273$ pneumonia images and $1583$ normal images of the chest.   
	\item \textbf{PACS dataset} \cite{pacspaper}: We use the dataset to study calibration under domain shift. The dataset comprises of a total $9991$ images spread across $4$ different domains with $7$ classes each. The domains are namely Photo, Art, Sketch and Cartoon. We fine-tune the ResNet-18 model, pretrained ImageNet dataset, on the Photo domain using various competing techniques and test on the other three domains to measure how calibration holds in a domain shift. Following \cite{pacspaper}, we also divide the training set of photo domain into $9:1$ train/val set.
	\item \textbf{Rotated MNIST Dataset}: This dataset is also used for domain shift experiments. Inspired from \cite{domain-drift-calibration-posthoc}, we create 5 different test sets namely $\{M_{15}, M_{30}, M_{45}, M_{60}, M_{75}\}$. Domain drift is introduced in each $M_{x}$ by rotating the images in the MNIST test set by $x$ degrees counter-clockwise. 
	
	\item \textbf{Segmentation Datasets- PASCAL VOC 2012 \cite{pascal-voc-2012}}: This a segmentation dataset and consists of images with pixel-level annotations. There are 21 classes overall (One background class and 20 foreground object classes). The dataset is divided into \textit{Train} (1464 images), \textit{Val} (1449 images) and \textit{Test} (1456 images) set. We, however, only make use of the \textit{Train} and the \textit{Val} set. Models are trained on the \textit{Train} set and the evaluation is reported on the \textit{Val} set. 
\end{enumerate}

%\textbf{(b) Kather 5000} \cite{katherpaper} : Kather 5000, as the name suggests, contains 5000 RGB images having 150x150 dimensions. Each image belong to exactly one of the 8 following listed tissues:  tumour epithelium, simple stroma, complex stroma, immune cells, debris, normal mucosal glands, adipose tissue, background (no tissue). We randomly sample a train/test split having a ratio of 4:1.

\section{Training Procedure and Implementation}
\label{sec:trainProc}

\myfirstpara{Backbone Architecture}
We use ResNet \cite{resnetpaper} backbone for most of our experiments. For training on CIFAR10, and CIFAR100 datasets we used ResNet-32 and ResNet-56. For SVHN we used ResNet-20 and ResNet-56. Following \cite{dcapaper}, for Mendeley V2 dataset, we use a ResNet-50 architecture pre-trained on ImageNet dataset \cite{imagenetpaper}. We use the backbone as a fixed feature extractor and add a $1 \times 1$ convolutional layer and two fully connected layers on top of the feature extractor. Segmentation experiments make use of DeepLabV3+ \cite{deeplabv3Plus} based on the Xception65 \cite{xception} backbone

% Kather 5000 is also trained similar to Mendeley V2, however, we use a ResNet-34 (pre-trained on ImageNet \cite{imagenetpaper}) as a feature extractor instead of ResNet-56.

\mypara{Train Parameters}
For all our experiments we used a single Nvidia 1080 Ti GPU. We trained CIFAR10 for a total of $160$ epochs using a learning rate of $0.1$, and it is reduced by a factor of $10$ at the $80^{th}$ and $160^{th}$ epoch. Train batch size was kept at $128$ and DNN was optimized using Stochastic Gradient Descent (SGD) with momentum at $0.9$ and weight decay being $0.0005$. Furthermore, we augment the images using random center crop and horizontal flips. We use same parameters for CIFAR100, except the number of epochs, where we train it for $200$ epochs. Again learning rate is reduced by a factor of $10$, but this time it is reduced at epochs 100 and 150. For Tiny-ImageNet, we followed the same training procedure as done by \cite{focallosspaper}. For SVHN, we keep the same training procedure as above except the number of epochs; we train for $100$ epochs, with it getting reduced at epochs $50$ and $70$ with a factor of $10$ yet again. We do not augment the images when training on SVHN. We use PyTorch framework for all our implementations. Our repository is inspired from \url{https://github.com/bearpaw/pytorch-classification}. We also take help from the official implementation of \cite{focallosspaper} to implement some of the baseline methods. For the segmentation experiments we make use of the following repository: \url{https://github.com/LikeLy-Journey/SegmenTron}. We train the segmentation models for 120 epochs using SGD optimizer with a warm-up LR scheduler. To train with focal loss, we used $\gamma = 3$. The rest of the parameters for the optimizer and the scheduler were kept the same as provided by the SegmenTron repository. 

\begin{figure}
	\centering
	\includegraphics[width =0.45\textwidth]{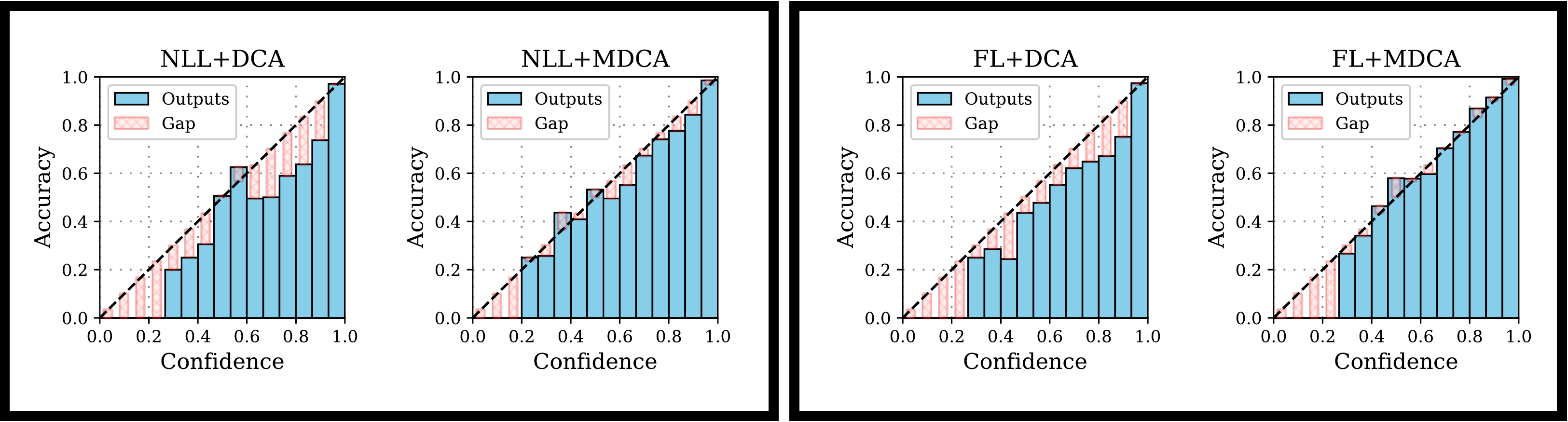}
	\caption{Comparison of Reliability diagrams of: (left) \texttt{NLL+DCA} vs \texttt{NLL+MDCA} and (right) \texttt{FL+DCA} vs. \texttt{FL+MDCA}. We use ResNet-32 trained on CIFAR10 dataset for comparison.}
	\label{fig:DCAvsMDCA}
\end{figure}

\begin{figure}
	\centering
	\includegraphics[width =0.45\textwidth]{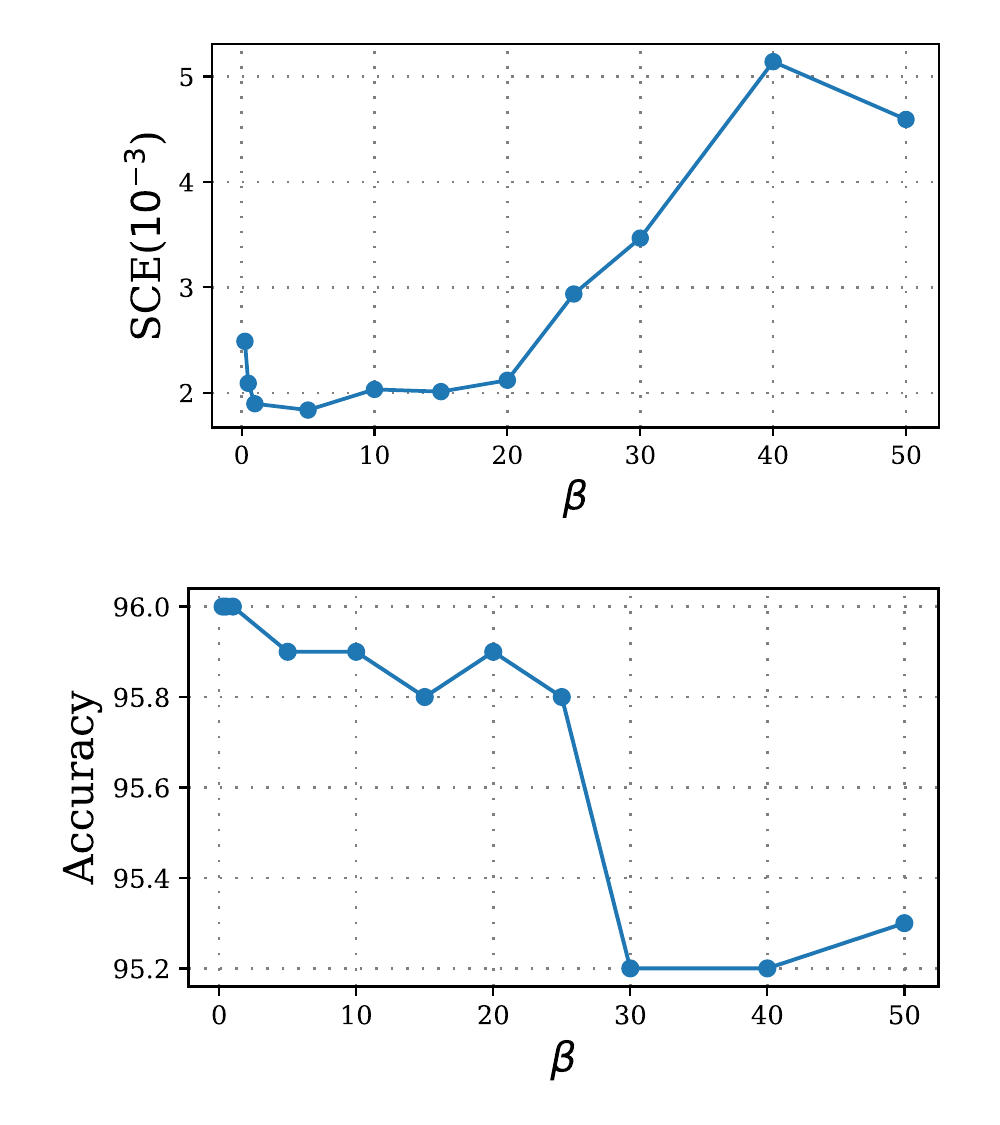}
	\caption{Effect of $\beta$ on Accuracy and SCE ($10^{-3}$) on \texttt{FLL+MDCA} on the ResNet-20 model trained on SVHN dataset.}
	\label{fig:EffectofbetaOnSCE}
\end{figure}

\mypara{Optimizing Hyper-parameter $\beta$ for MDCA}
We vary the hyper-parameter $\beta \in [0.25, 0.5, 1,5,10, \dots 50]$ in our experiments. Figure \ref{fig:EffectofbetaOnSCE} shows how calibration error and accuracy is affected when we increase $\beta$ in different model-dataset pairs. We see a general trend that calibration is best achieved when $\beta$ is close to $1$, and as we increase it, the calibration as well as the accuracy starts to drop (accuracy decreases and the SCE score increases). 

%\chetan{Why is this important. How is it related to differentiability of MDCA. Delete if we are not able to answer the above two points.}  However, unlike DCA, which did not train with higher $\beta$ values ($\beta > 10$), a model is still able to train with MDCA for high $\beta$ values. This is due to the differentiability of our loss function.

\mypara{Post-Hoc Calibration Experiments} 
For comparison with post-hoc techniques, we set aside $10\%$ of the training data as a validation set (hold-out set) to conduct post-hoc calibration. For Temperature Scaling (\texttt{TS}), we perform a grid search between the range of $0$ to $10$ with a step-size of $0.1$ to find the optimal temperature value that gives the least \texttt{NLL} on the hold-out set. For Dirichlet Calibration (\texttt{DC}), we attach a single layer neural network at the end of the DNN and use ODIR \cite{Dirichlet} regularization on the weights of the same. We train on the hold-out set keeping the rest of the DNN weights frozen except the newly added layer. We again use grid search to find the optimal hyper-parameters $\lambda$ and $\mu$ that give the least \texttt{NLL} on the hold-out set. We vary $\lambda \in \{ 0, 0.01, 0.1, 1, 10, 0.005, 0.05, 0.5, 5, 0.0025, 0.025, 0.25, 2.5 \}$ and $\mu \in \{0, 0.01, 0.1, 1, 10\}$.

\begin{table}[t]
	\resizebox{1.0 \linewidth}{!}{%
		\begin{tabular}{lcccc}
			\toprule
			\textbf{$\mathcal{L}_{C}$}          & \textbf{$\mathcal{L}_{Auxiliary}$} & \textbf{Accuracy} & \textbf{SCE ($10^{-3}$)} & \textbf{ECE ($\%$)} \\ \midrule
			\textbf{ResNet-32 on CIFAR10 dataset:}\\
			\bottomrule\\
			\multirow{4}{*}{Cross Entropy} & None          & 92.54        & 8.68                                    & 4.25            \\
			& DCA \cite{dcapaper}           & {92.94}        & 8.41                                    & 4.00            \\
			& MMCE \cite{kumarpaper}          & 91.59        & 8.17                                    & 3.31            \\
			& MDCA (ours)          & 91.85        & {4.63}                                    & {1.69}            \\ \midrule
			\multirow{4}{*}{Label Smoothing \cite{labelsmoothinghelp}}  & None          & 92.48        & 14.09                                   & 6.28            \\
			& DCA \cite{dcapaper}           & 88.69        & {8.14}                                    & {3.44}            \\
			& MMCE \cite{kumarpaper}          & 91.92        & 22.49                                   & 10.07           \\
			& MDCA (ours)          & {93.01}        & 12.38                                   & 5.66            \\ \midrule
			\multirow{4}{*}{Focal Loss \cite{ogfocalloss}}  & None          & 92.52        & 4.09                                    & 1.01            \\
			& DCA \cite{dcapaper}           & 92.65        & 7.50                                    & 3.49            \\
			& MMCE \cite{kumarpaper}          & 90.76        & 26.41                                   & 13.38           \\
			& MDCA (ours)          & {92.82}        & {3.22}                                    & {0.93}       \\ 
			\bottomrule\\
			\textbf{ResNet-20 on SVHN dataset:}\\
			\bottomrule\\
			\multirow{4}{*}{Cross Entropy} & None          & 96.14        & 3.43                                    & 1.64            \\
			& DCA \cite{dcapaper}           & 96.17        & 4.29                                    & 2.02            \\
			& MMCE \cite{kumarpaper}          & 95.88        & 9.18                                    & 4.34            \\
			& MDCA (ours)         & {96.33}        & {1.46}                                    & {0.43}            \\ \midrule
			\multirow{4}{*}{Label Smoothing \cite{labelsmoothinghelp}}  & None          & 96.24        & 18.80                                   & 8.88            \\
			& DCA \cite{dcapaper}           & {96.55}        & {9.80}                                    & {4.46}            \\
			& MMCE \cite{kumarpaper}          & 96.11        & 31.75                                   & 15.50           \\
			& MDCA (ours)          & 96.14        & 13.91                                   & 6.46            \\ \midrule
			\multirow{4}{*}{Focal Loss \cite{ogfocalloss}}  & None          & 96.17        & 2.54                                    & 0.89            \\
			& DCA \cite{dcapaper}           & {96.35}        & 2.12                                    & {0.44}            \\
			& MMCE \cite{kumarpaper}          & 95.83        & 25.01                                   & 12.79           \\
			& MDCA (ours)          & 96.08        & {1.90}                                    & 0.47      \\ \bottomrule     
		\end{tabular}%
	}
	\caption{Ablation to study the impact of various auxiliary losses on two models : ResNet-32 on CIFAR10 and Resnet-20 trained model on SVHN Dataset. Our approach exhibits least calibration error without sacrificing on the accuracy.}
	\label{tab:AblCIFARnSVHN}
\end{table}
\begin{table}[t]
	\centering
	\resizebox{0.9\linewidth}{!}{%
		\begin{tabular}{l|ccc|c}
			\toprule
			\textbf{Methods}        & \textbf{Art} & \textbf{Cartoon} & \textbf{Sketch} & \textbf{Average} \\ \midrule
			\midrule
			\multicolumn{5}{c}{\textbf{SCE($10^{-3})$}} \\
			\midrule \midrule
			NLL                                 & 6.33          & 17.95          & 15.01          & 13.10              \\
			LS \cite{labelsmoothinghelp}                                  & 7.80          & 11.95          & \textbf{10.88} & 10.21              \\
			FL \cite{ogfocalloss}                                  & 8.61          & 16.62          & 10.94          & 12.06              \\
			Brier Score \cite{brierloss}                               & 6.55          & 13.19          & 15.63          & 11.79              \\
			MMCE \cite{kumarpaper}                                & 6.35          & 15.70          & 17.16          & 13.07              \\
			DCA \cite{dcapaper}                                 & 7.49          & 18.01          & 14.99          & 13.49              \\ 
			FLSD \cite{focallosspaper}                       & 8.35          & 13.39          & 13.86          & 11.87              \\
			\midrule
			%\rowcolor{LightCyan}
			%NLL+MDCA                           & \textbf{5.10} & 16.53          & 13.49          & 11.71              \\
			%\rowcolor{LightCyan}
			%LS+MDCA                            & 5.70          & 12.07          & 11.70          & 9.82               \\
			\rowcolor{LightCyan}
			Ours (FL+MDCA)                          & \textbf{6.21}          & \textbf{11.91} & 11.08          & \textbf{9.73} \\
			\midrule \midrule

			\multicolumn{5}{c}{\textbf{ECE (\%)}}\\
			\midrule \midrule
			%\textbf{Methods}        & \textbf{Art} & \textbf{Cartoon} & \textbf{Sketch} & \textbf{Average} \\ \midrule
			Brier Score~\cite{brierloss}    & 3.35         & 33.68            & 42.61           & 26.55            \\
			NLL            & 9.42         & 52.99            & 35.56           & 32.66            \\
			LS~\cite{labelsmoothinghelp}             & 8.70         & 25.21            & 13.29           & 15.73            \\
			Focal Loss~\cite{ogfocalloss}     & 7.34         & 48.96            & 25.33           & 27.21            \\
			MMCE~\cite{kumarpaper}           & 17.06        & 43.25            & 40.79           & 33.70            \\
			DCA~\cite{dcapaper}            & 13.38        & 55.20            & 37.76           & 35.45            \\
			FLSD~\cite{focallosspaper}           & 8.41         & 34.43            & 30.01           & 24.28            \\ \midrule
			
			\textbf{Ours (FL+MDCA)} & 6.29         & 29.81            & 23.05           & 19.71  \\ 
			\midrule \midrule   
			
			\multicolumn{5}{c}{\textbf{Top-1 (\%) Accuracy}} \\
			\midrule \midrule
			Brier Score\cite{brierloss}    & 59.28        & 24.83            & 28.43           & 37.51            \\
			NLL            & 59.08        & 21.20            & 28.00           & 36.09            \\
			LS\cite{labelsmoothinghelp}             & 56.35        & 22.01            & 29.88           & 36.08            \\
			FL\cite{ogfocalloss}     & 52.83        & 17.32            & 26.70           & 32.28            \\
			MMCE\cite{kumarpaper}           & 60.60        & 29.95            & 20.36           & 36.97            \\
			DCA\cite{dcapaper}            & 57.91        & 20.86            & 28.30           & 35.69            \\
			FLSD\cite{focallosspaper}           & 54.54        & 20.82            & 29.35           & 34.90            \\ \midrule
			\textbf{Ours (FL+MDCA)} & 63.23        & 27.86            & 23.01           & 38.03      \\ \bottomrule     
			
		\end{tabular}%
	}
	\caption{Table comparing SCE, ECE and Top-1\% Accuracy values of our method against others on PACS~\cite{pacspaper} dataset when the model is trained on \texttt{Photo} domain and tested on domains: \texttt{Art, Cartoon, and Sketch}}
	\label{tab:rmnist-pacs-SCE-ece-top1-all_methods}
\end{table}

\begin{table}[]
\resizebox{0.5 \textwidth}{!}{%
\begin{tabular}{lcccccc|c}
\toprule
\multicolumn{1}{c}{\textbf{Method}} & {Clean}      & \textbf{$M_{15}$}     & \textbf{$M_{30}$}     & \textbf{$M_{45}$}     & \textbf{$M_{60}$}     & \textbf{$M_{75}$}     & \textbf{Average} \\ \midrule \midrule 
\multicolumn{8}{c}{\textbf{SCE ($\%$)}} \\ \midrule \midrule
NLL             & {0.07} &  {0.19} & 0.70          & 2.96          & 7.50          & 10.60         & 3.67           \\
LS \cite{labelsmoothinghelp}              & 2.00          & 2.00          & 1.93          & 2.91          & 6.67          & 8.93          & 4.07           \\
FL \cite{ogfocalloss}              & 0.29          & 0.81          & 1.34          & 2.62          & 7.04          & 10.10         & 3.70           \\
Brier Score \cite{brierloss}           & 0.23          & 0.51          & 1.09          & 2.83          & 6.58          & 9.10          & 3.39           \\
MMCE \cite{kumarpaper}            & 2.51          & 4.05          & 5.01          & 4.55          & {5.28} & 5.29 & 4.45           \\
DCA \cite{dcapaper}             & 0.07               & 0.20               & 0.91               & 3.71               & 8.42               & 11.65               & 4.16                \\
FLSD \cite{focallosspaper}      & 1.30               & 2.09               & 3.10               & 3.05               & 4.88               & 7.56               & 3.67                \\ 
\midrule
\textbf{Ours (FL+MDCA)}                                    & 0.20          & 0.48          & 0.94          & 2.51 & 6.65          & 9.61          & {3.40} \\ \midrule \midrule 
\multicolumn{8}{c}{\textbf{ECE ($\%$)}} \\ \midrule \midrule

NLL                                & 0.11 & 0.38  & 1.13  & 10.54 & 31.95 & 45.06 & 14.86 \\
LS \cite{labelsmoothinghelp}          & 9.89 & 9.32  & 7.13  & 6.86  & 27.36 & 37.28 & 16.31 \\
FL \cite{ogfocalloss}         & 1.34 & 3.11  & 3.92  & 5.21  & 28.11 & 41.17 & 13.81 \\
Brier Score \cite{brierloss}          & 0.90 & 2.07  & 2.57  & 5.95  & 25.24 & 38.16 & 12.48 \\
MMCE \cite{kumarpaper}                & 4.72 & 15.42 & 23.01 & 18.88 & 10.74 & 6.80  & 13.26 \\
DCA \cite{dcapaper}                   & 0.21 & 0.29  & 1.88  & 13.83 & 36.31 & 50.25 & 17.13 \\
FLSD \cite{focallosspaper}            & 6.50 & 10.43 & 14.33 & 7.92  & 14.80 & 29.34 & 13.89 \\ \midrule
\textbf{Ours (FL+MDCA)}                                      & 0.79 & 2.03  & 2.94  & 6.18  & 25.77 & 39.29 & 12.83 \\ \midrule \midrule 
\multicolumn{8}{c}{\textbf{Top-1 (\%) Accuracy}} \\ \midrule \midrule

NLL                                & 99.61 & 98.79 & 93.38 & 73.82 & 43.24 & 24.07 &  72.15 \\
LS \cite{labelsmoothinghelp}          & 99.62 & 98.39 & 92.04 & 69.33 & 39.94 & 22.99 &  70.39 \\
FL \cite{ogfocalloss}         & 99.63 & 98.20 & 91.87 & 69.95 & 38.67 & 20.83 &  69.86 \\
Brier Score \cite{brierloss}          & 99.63 & 98.72 & 92.54 & 71.29 & 41.90 & 22.73 &  71.14 \\
MMCE \cite{kumarpaper}                & 98.43 & 94.99 & 83.93 & 53.88 & 28.76 & 16.84 &  62.81 \\
DCA \cite{dcapaper}                   & 99.60 & 98.36 & 91.51 & 68.34 & 38.32 & 20.93 &  69.51 \\
FLSD \cite{focallosspaper}            & 99.67 & 98.79 & 92.77 & 71.17 & 40.17 & 20.84 &  70.57 \\ \midrule
\textbf{Ours (FL+MDCA)}                                     & 99.59 & 98.61 & 93.13 & 70.92 & 41.93 & 23.37 &  71.26 \\ \bottomrule

\end{tabular}%
}
\caption{Table comparing SCE, ECE and Top-1\% Accuracy  values of our loss with other methods on the Rotated-MNIST dataset trained using a ResNet-20 model.}
\label{tab:SCE-rot-MNIST}
\end{table}

\mypara{Domain Shift Experiments} 
For PACS dataset, we use the official PyTorch ResNet-18 model pre-trained on ImageNet dataset. We re-initialized its last fully connected layer to accommodate 7 classes, and finally fine-tuned on the Photo domain. We use the SGD optimizer with same momentum and weight decay values as done for CIFAR10/100 and described earlier. Training batch size was fixed at $256$, and the model was trained for 30 epochs with initial learning rate, set at 0.01, getting reduced at epoch 20 with a factor of 10. Training parameters are chosen such that they give the best performing model i.e. having the maximum accuracy on the Photo domain val set.

For Rotated MNIST, we used PyTorch framework to generate rotated images. We use the ResNet-20 model to train on the standard MNIST train set for 30 epochs with a learning rate of 0.1 for first 20 epochs, and then with 0.01 for the last 10 epochs. Rest of the details like batch size and optimizer remain same as the CIFAR10/100 experiments. We did not augment any images, and selected the training parameters such that the model gives best accuracy on the validation set.

For 20 Newsgroups, we train the Global Pooling Convolutional Network \cite{global-pool-cnn} using the ADAM optimizer, with learning rate $0.001$, and the default values of betas at $0.9$ and $0.999$ respectively. We used \texttt{GloVe} word embeddings \cite{glove-vectors} to train the network. We trained the model for $50$ epochs.

\section{Additional Results}
\label{sec:addnlRes}

We report the following additional results:

\begin{table}[t]
\centering
\resizebox{0.9\linewidth}{!}{%
\begin{tabular}{l|cc|cc|cc}
\toprule
\textbf{Domain}  & \textbf{NLL} & \textbf{NLL+MDCA} & \textbf{LS} & \textbf{LS+MDCA} & \textbf{FL} & \textbf{FL+MDCA} \\ 
\midrule \midrule
\multicolumn{7}{c}{\textbf{SCE ($\%$)}}\\   
\midrule \midrule
			Clean                          & 0.07           & {0.07}     & 2.00      & {1.99}    & 0.29      & {0.20}    \\
			$M_{15}$                         & 0.19           & {0.19}     & 2.00      & {1.99}    & 0.81      & {0.48}    \\
			$M_{30}$                         & 0.70           & {0.66}     & 1.93      & {1.88}    & 1.34      & {0.94}    \\
			$M_{45}$                         & {2.96}  & 3.17              & 2.91      & {2.85}    & 2.62      & {2.51}    \\
			$M_{60}$                         & {7.50}  & 7.99              & 6.67      & {6.08}    & 7.04      & {6.65}    \\
			$M_{75}$                         & {10.60} & 11.17             & 8.93      & {8.82}    & 10.10     & {9.61}    \\ \midrule
			\rowcolor{LightCyan}
			\textbf{Average}                    & \textbf{3.67}  & 3.88              & 4.07      & \textbf{3.94}    & 3.70      & \textbf{3.40}  \\ \midrule 
\midrule
\multicolumn{7}{c}{\textbf{ECE (\%)}}\\
\midrule \midrule
Clean            & 0.11         & 0.22              & 9.89        & 9.83             & 1.34        & 0.79             \\
$M_{15}$         & 0.38         & 0.21              & 9.32        & 9.41             & 3.11        & 2.03             \\
$M_{30}$         & 1.13         & 1.31              & 7.13        & 7.38             & 3.92        & 2.94             \\
$M_{45}$         & 10.54        & 12.50             & 6.86        & 8.41             & 5.21        & 6.18             \\
$M_{60}$         & 31.95        & 35.02             & 27.36       & 25.32            & 28.11       & 25.77            \\
$M_{75}$         & 45.06        & 49.41             & 37.28       & 38.41            & 41.17       & 39.29            \\ \midrule
\textbf{Average} & 14.86        & 16.45             & 16.31       & 16.46            & 13.81       & 12.83 \\ 
\midrule \midrule
\multicolumn{7}{c}{\textbf{Top-1 (\%)} Accuracy}\\    
\midrule \midrule
Clean            & 99.61                            & 99.64                                  & 99.62                           & 99.63                                 & 99.63                           & 99.59                                \\
$M_{15}$         & 98.79                            & 98.63                                  & 98.39                           & 98.45                                 & 98.20                           & 98.61                                \\
$M_{30}$         & 93.38                            & 93.23                                  & 92.04                           & 92.54                                 & 91.87                           & 93.13                                \\
$M_{45}$         & 73.82                            & 71.53                                  & 69.33                           & 69.74                                 & 69.95                           & 70.92                                \\
$M_{60}$         & 43.24                            & 41.27                                  & 39.94                           & 41.88                                 & 38.67                           & 41.93                                \\
$M_{75}$         & 24.07                            & 22.05                                  & 22.99                           & 22.97                                 & 20.83                           & 23.37                                \\ \midrule
\textbf{Average} & 72.15                            & 71.06                                  & 70.39                           & 70.87                                 & 69.86                           & 71.26    \\ \bottomrule
\end{tabular}%
}
\caption{Table comparing SCE, ECE and Top-1\% Accuracy values of our method against others for ResNet-20 model on Rotated-MNIST dataset. \emph{Clean} denotes the original dataset, and subscript under the `M' indicates angle of rotation for each digit.}
\label{tab:rot-mnist-ece-top1-ablation}
\end{table}

\begin{table}[t]
	\centering
	\resizebox{\linewidth}{!}{%
		\begin{tabular}{c|cc|cc|cc}
			\toprule
			\textbf{Domain}  & \textbf{NLL} & \textbf{NLL+MDCA} & \textbf{LS} & \textbf{LS+MDCA} & \textbf{FL} & \textbf{FL+MDCA} \\ \midrule \midrule
			\multicolumn{7}{c}{\textbf{SCE ($10^{-3}$) }}\\
			% \midrule
			% 			NLL                                 & 6.33          & 17.95          & 15.01          & 13.10              \\
			% 			LS \citeyearpar{labelsmoothinghelp}                                  & 7.80          & 11.95          & \textbf{10.88} & 10.21              \\
			% 			FL \citeyearpar{ogfocalloss}                                  & 8.61          & 16.62          & 10.94          & 12.06              \\
			% 			Brier Score \citeyearpar{brierloss}                               & 6.55          & 13.19          & 15.63          & 11.79              \\
			% 			MMCE \citeyearpar{kumarpaper}                                & 6.35          & 15.70          & 17.16          & 13.07              \\
			% 			DCA \citeyearpar{dcapaper}                                 & 7.49          & 18.01          & 14.99          & 13.49              \\ 
			% 			FLSD \citeyearpar{focallosspaper}                       & 8.35          & 13.39          & 13.86          & 11.87              \\
			% 			\midrule
			% 			%\rowcolor{LightCyan}
			% 			%NLL+MDCA                           & \textbf{5.10} & 16.53          & 13.49          & 11.71              \\
			% 			%\rowcolor{LightCyan}
			% 			%LS+MDCA                            & 5.70          & 12.07          & 11.70          & 9.82               \\
			% 			\rowcolor{LightCyan}
			% 			Ours (FL+MDCA)                          & \textbf{6.21}          & \textbf{11.91} & 11.08          & \textbf{9.73} \\
			\midrule \midrule
			\textbf{Art}     & 6.33         & 5.10             & 7.80        & 5.70            & 8.61        & 6.21             \\
			\textbf{Cartoon} & 17.95        & 16.53             & 11.95       & 12.07            & 16.22       & 11.91            \\
			\textbf{Sketch}  & 15.01       & 13.49            & 10.88       & 11.70            & 10.94       & 11.08            \\ \midrule
			\rowcolor{LightCyan}
			\textbf{Average} & 13.10        & \textbf{11.71 }           & 10.21       & \textbf{9.82}            & 12.06       & \textbf{ 9.73}     \\ \midrule  \midrule
			%\rowcolor{LightCyan}
			\multicolumn{7}{c}{\textbf{ECE (\%) }}\\
			\midrule
			\midrule
			\textbf{Art}     & 9.42         & 4.99              & 8.70        & 11.80            & 7.34        & 6.29             \\
			\textbf{Cartoon} & 52.99        & 47.96             & 25.21       & 22.73            & 48.96       & 29.81            \\
			\textbf{Sketch}  & 35.56        & 31.68             & 13.29       & 10.34            & 25.33       & 23.05            \\ \midrule
			\rowcolor{LightCyan}
			\textbf{Average} & 32.66        & \textbf{28.21}             & 15.73       & \textbf{14.96}            & 27.21       & \textbf{19.71}     \\ \midrule   \midrule
			\multicolumn{7}{c}{\textbf{Top-1 (\%) Accuracy }}\\
			\midrule \midrule
			\textbf{Art}     & 59.08                            & 61.67                                  & 56.35                           & 59.57                                 & 52.83                           & 63.23                                \\
			\textbf{Cartoon} & 21.20                            & 21.72                                  & 22.01                           & 24.62                                 & 17.32                           & 27.86                                \\
			\textbf{Sketch}  & 28.00                            & 26.90                                  & 29.88                           & 26.55                                 & 26.70                           & 23.01                                \\ \midrule
			\rowcolor{LightCyan}
			\textbf{Average} & 36.09                            & \textbf{36.76}                                  & 36.08                           & \textbf{36.91}                                 & 32.28                           & \textbf{ 38.03}     \\ \bottomrule                          
			
		\end{tabular}%
	}
	\caption{Ablation comparing SCE, ECE and Top-1 Accuracy values of models using \texttt{MDCA} as an auxiliary loss along with other classification losses. The numbers correspond to training a ResNet-18 model on \texttt{Photo} subset from PACS dataset, and testing on other subsets of the PACS.}
	\label{tab:ece-top1-rot-pacs-ablation}
\end{table}

\begin{figure}[t]
	\centering
	\includegraphics[width =\linewidth]{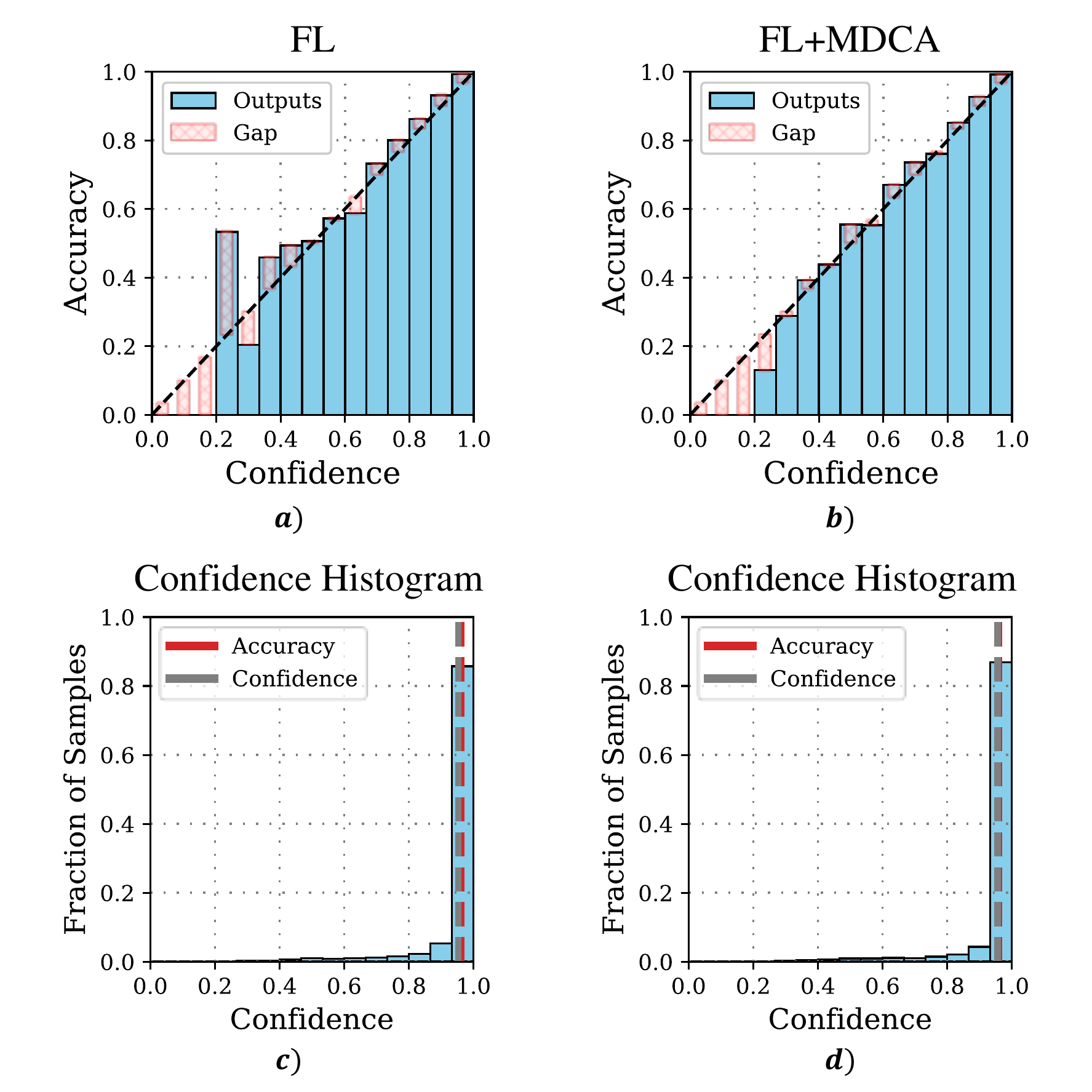}
	%\vspace{-4 mm}
	\caption{Reliability diagrams (a,b) and confidence histograms (c,d) of \texttt{FL} trained model compared against MDCA regularized version (\texttt{FL+MDCA}). We use ResNet-32 trained on CIFAR10 dataset for comparison.}
	\label{fig:RelDigFLMDCA_vs_FL}
\end{figure}

\begin{enumerate}[label*=\arabic*.]
	\item \textbf{Class-j-ECE score:} In Table 3 of main manuscript we reported the Class-j-ECE score for SVHN. In Table \ref{tab:classECE_CIFAR10} here, we we provide additional results for CIFAR10. 
	% \cref{tab:sel-config}
	\item \textbf{Comparison with other auxiliary losses}: In  Table 6 in the main manuscript showed how the proposed MDCA can be used along with \texttt{NLL}, \texttt{LS} \cite{originallabelsmoothing}, and \texttt{FL} \cite{ogfocalloss} to improve the calibration performance without sacrificing the accuracy. In Tab. \ref{tab:AblCIFARnSVHN} here, we show a similar comparison for other competitive approaches, namely \texttt{DCA} \cite{dcapaper}, and \texttt{MMCE} \cite{kumarpaper}. Using \texttt{MDCA}, gives better calibration than other competitive approaches.
	% \cref{tab:datasetdrift}
	\item \textbf{Calibration performance under dataset drift:} A model trained using our proposed loss gives better calibration under dataset drift as well. Table 4 in the main manuscript showed SCE score comparison on PACS. We give more detailed comparison here in  Tab. \ref{tab:rmnist-pacs-SCE-ece-top1-all_methods} which shows top 1\% accuracy, ECE as well. We repeat SCE numbers from main manuscript for completion.
	Tab. \ref{tab:SCE-rot-MNIST} shows the corresponding numbers for Rotated MNIST.
	
	Just like we showed that using \texttt{MDCA} in conjunction with \texttt{NLL}, \texttt{LS} \cite{originallabelsmoothing}, and \texttt{FL} \cite{ogfocalloss} gives best calibration performance, we show that this remains true even for the dataset drift case. Tab. \ref{tab:rot-mnist-ece-top1-ablation} and Tab. \ref{tab:ece-top1-rot-pacs-ablation} show the comparison on Rotated MNIST and PACS datasets respectively. 
	%Similarly the SCE scores for Rotated MINST have been reported in \cref{tab:SCE-rot-MNIST}. To conclusively understand the role of MDCA regulariser in conjunction with other application specific losses, refer to Table \ref{tab:ece-top1-rot-pacs-ablation} and \cref{tab:rot-mnist-ece-top1-ablation}. 
	% \ref{fig:relDiagramNLL} and \ref{fig:relDiagramLS}
	\item \textbf{Reliability Diagrams:} Fig. 2 in the main manuscript showed reliability and confidence plots for \texttt{MDCA} used with \texttt{NLL} and \texttt{LS} respectively. We show similar plots for \texttt{MDCA+FL} in Fig. \ref{fig:RelDigFLMDCA_vs_FL}.	
	%\cite{reliabilitydiagram} are a common method to visualise calibration; plots accuracies and confidence bins as a bar chart. When a DNN is perfectly calibrated, the accuracy for each bin would be equal to that of the confidence of that particular bin, therefore all the bars would lie on $y=x$ curve. If bars are below the $y=x$ curve, then the model is less accurate than it is confident i.e., deemed over-confident. If the bars are above the $y=x$ curve, this means that the DNN is less confident than it is accurate, is considered under-confident. For a multi-class setting, we can also plot a \textit{Classwise-reliability diagram} \cite{Dirichlet} wherein for each class, we plot a reliability diagram as if it was only the class the DNN predicts. \cref{fig:RelDigFLMDCA_vs_FL} shows a representative reliability diagram obtained by \texttt{FL+MDCA} with \texttt{FL} considered in isolation. 
	%
\end{enumerate}

\begin{table}[t]
	\resizebox{1.0\linewidth}{!}{%
		\begin{tabular}{lcccccccccc}
			\toprule
			\multicolumn{1}{c}{\multirow{2}{*}{\textbf{Method}}} & \multicolumn{10}{c}{\textbf{Classes}}                                                                            \\
			\multicolumn{1}{c}{}                                 & \multicolumn{1}{c}{\textbf{airplane}} & \multicolumn{1}{c}{\textbf{automobile}} & \multicolumn{1}{c}{\textbf{bird}} & \multicolumn{1}{c}{\textbf{cat}} & \multicolumn{1}{c}{\textbf{deer}} & \multicolumn{1}{c}{\textbf{dog}} & \multicolumn{1}{c}{\textbf{frog}} & \multicolumn{1}{c}{\textbf{horse}} & \multicolumn{1}{c}{\textbf{ship}} & \multicolumn{1}{c}{\textbf{truck}} \\ \midrule
			Cross Entropy                                                  & 0.98                           & 0.43                           & 1.12                           & 1.97                           & 0.65                           & 1.47                           & 0.65                           & 0.44                           & 0.58                           & 0.57                           \\
			Focal Loss \cite{ogfocalloss}                                   & 0.38                           & \textbf{0.23}                  & 0.69                           & 1.08                           & 0.39                  & 0.91                           & 0.37                  & \textbf{0.34}                           & 0.25                  & \textbf{0.24}                  \\
			LS \cite{labelsmoothinghelp}                            & 1.64                           & 1.89                           & 1.26                           & 1.01                  & 1.64                           & 1.25                           & 1.66                           & 1.62                           & 1.76                           & 1.77                           \\
			Brier Score \cite{brierloss}                                         & 0.71                           & 0.25                           & 0.91                           & 1.56                           & 0.61                           & 1.26                           & 0.4                            & 0.34                           & 0.41                           & 0.37                           \\
			MMCE \cite{kumarpaper}                                                 & 1.88                           & 1.29                           & 1.57                           & 2.43                           & 1.83                           & 1.62                           & 1.57                           & 1.51                           & 1.09                           & 1.75                           \\
			DCA \cite{dcapaper}                                   & 0.80                            & 0.43                           & 1.18                           & 1.71                           & 0.93                           & 1.44                           & 0.52                           & 0.55                           & 0.51                           & 0.6     \\ 
			FLSD \cite{focallosspaper} & 0.99 & 1.12 & 0.81 & 1.11 & 0.81 & 1.44 & 0.81 & 0.85 & 0.70 & 1.14 \\ \midrule
			% NLL+MDCA (ours)                                      & 0.54                  & 0.30                            & 0.53                  & 1.09                           & 0.46                           & 0.61                  & 0.55                           & \textbf{0.28}                  & 0.33                           & 0.38                           \\
			% LS+MDCA (ours)                          & 0.91                           & 1.43                           & 0.93                           & 0.95                           & 1.24                           & 1.23                           & 1.21                           & 1.25                           & 1.3                            & 1.31\\
			\rowcolor{LightCyan}
			\textbf{Ours (FL+MDCA)}          & \textbf{0.36} & 0.37 & \textbf{0.36} & \textbf{0.60} & \textbf{0.35} & \textbf{0.59} & \textbf{0.31} & 0.41 & \textbf{0.25} & 0.42  \\
			\bottomrule        
			
		\end{tabular}%
		
	}
	
	\caption{Class-$j$-ECE (\%) values on all ten classes for a ResNet-32 model trained on the CIFAR10 dataset comparing different learnable calibration methods including ours highlighted in Cyan.}
	\label{tab:classECE_CIFAR10}
\end{table}

\clearpage

\newpage
{\small
\bibliographystyle{ieee_fullname}
\bibliography{egbib}

}

\end{document}